\documentclass[preprint,12pt]{elsarticle}

\usepackage{amsmath,amssymb,amsfonts}
\usepackage{graphicx}
\usepackage{textcomp}
\usepackage{xcolor}
\def\BibTeX{{\rm B\kern-.05em{\sc i\kern-.025em b}\kern-.08em
    T\kern-.1667em\lower.7ex\hbox{E}\kern-.125emX}}
\usepackage{algorithm2e}
\makeatletter
\renewcommand{\@algocf@capt@plain}{above}
\makeatother
\usepackage{caption}
\usepackage{multicol,lipsum}
\usepackage{cuted}
\usepackage{mathtools}
\usepackage{array}
\usepackage[T1]{fontenc}
\usepackage{multirow}
\usepackage{longtable}
\usepackage{booktabs}
\usepackage[utf8]{inputenc}
\usepackage{titlesec}
\usepackage{epstopdf}
\usepackage{enumitem}
\usepackage{caption,subcaption}

\usepackage{verbatim}
\usepackage{color,soul}

\include{pythonlisting}

\include{pythonlisting}
\def\BibTeX{{\rm B\kern-.05em{\sc i\kern-.025em b}\kern-.08em
    T\kern-.1667em\lower.7ex\hbox{E}\kern-.125emX}}

\journal{Information Sciences}

\begin{document}

\begin{frontmatter}



\title{Large-scale Multi-objective Feature Selection:\\ A  Multi-phase Search Space Shrinking Approach}


\author[label1]{Azam Asilian Bidgoli} 

\affiliation[label1]{organization={Wilfred Laurier University},
            city={Waterloo},
            state={Ontario},
            country={Canada}}
\author[label2]{Shahryar Rahnamayan} 

\affiliation[label2]{organization={Brock University},
            city={St. Catharines},
            state={Ontario},
            country={Canada}}
\begin{abstract}
Feature selection is a crucial step in machine learning, especially for high-dimensional datasets, where irrelevant and redundant features can degrade model performance and increase computational costs. This paper proposes a novel large-scale multi-objective evolutionary  algorithm based on the search space shrinking,  termed LMSSS, to tackle the challenges of  feature selection particularly as a sparse optimization problem.
The method includes a shrinking scheme to reduce dimensionality of the search space by eliminating irrelevant features before the main evolutionary process. This is achieved through a ranking-based filtering method that evaluates features based on their correlation with class labels and frequency in an initial, cost-effective evolutionary process. Additionally, a smart crossover scheme based on voting between parent solutions is introduced, giving higher weight to the parent with better classification accuracy. An intelligent mutation process is also designed to target features prematurely excluded from the population, ensuring they are evaluated in combination with other features.
These integrated techniques allow the evolutionary process to explore the search space more efficiently and effectively, addressing the sparse and high-dimensional nature of large-scale feature selection problems. The effectiveness of the proposed  algorithm is demonstrated through comprehensive experiments on 15 large-scale datasets, showcasing its potential to identify more accurate feature subsets compared to state-of-the-art large-scale feature selection algorithms. These results highlight LMSSS's capability to improve model performance and computational efficiency, setting a new benchmark in the field.\end{abstract}


\begin{highlights}
\item Proposes a Novel Shrinking Scheme: The paper introduces a large-scale multi-objective evolutionary algorithm (LMSSS) that effectively reduces the search space by eliminating irrelevant features before the evolutionary process, significantly improving both computational efficiency and feature selection accuracy.

\item Advanced Feature Ranking Technique: The method incorporates a ranking-based filtering approach, ensuring that only the most relevant features are selected for further processing, improving the overall classification accuracy.

\item Voting-Based Crossover Scheme: A novel voting-based crossover technique is employed to ensure the optimal combination of selected features, which enhances the search process and contributes to better model performance on high-dimensional datasets.
\item Comprehensive Experimental Validation: The proposed LMSSS algorithm outperforms state-of-the-art large-scale feature selection methods across 15 large-scale datasets, demonstrating its capability to improve model performance while reducing computational cost.

\end{highlights}


\end{frontmatter}



\section{Introduction}

Feature selection is a critical task in machine learning and data analysis, as it involves identifying the most relevant features for building efficient and accurate predictive models~\cite{dhal2022comprehensive}. The importance of feature selection stems from its ability to enhance model performance by removing redundant or irrelevant features, thus reducing overfitting, improving model interpretability, and decreasing computational cost \cite {9928423,asilian2021novel}. In real-world applications, datasets often contain a multitude of features, making it imperative to employ effective feature selection techniques. Accordingly, feature selection can be defined as a large-scale optimization problem to find a set of high-quality features for a specific machine-learning task such as classification \cite{liu2024feature,bidgoli2019novel}, clustering \cite{alelyani2018feature}, and segmentation \cite{sharif2020active}. Multi-objective feature selection \cite {9641743} has gained prominence because it simultaneously optimizes multiple criteria, such as maximizing accuracy while minimizing the number of selected features, leading to more balanced and practical set of features.
 Evolutionary algorithms (EAs)~\cite{agrawal2021metaheuristic,9371430} have emerged as a promising approach for multi-objective feature selection due to their inherent capability to explore large and complex search spaces efficiently~\cite{asilian2022bias,bidgoli2022evolutionary}. EAs mimic the process of natural evolution, employing mechanisms such as selection, crossover, and mutation to evolve solutions over successive generations. Their population-based nature allows for the parallel exploration of multiple regions in the search space, increasing the likelihood of discovering optimal or near-optimal solutions. 
Despite significant advancements in the application of evolutionary algorithms for feature selection, there remain challenges that necessitate further research and development. One major drawback is the large-scale search space, especially when dealing with high-dimensional datasets. As the number of features increases, the search space grows exponentially, leading to a combinatorial explosion that can overwhelm the search process \cite{song2020variable,tian2021evolutionary}. This vast search space makes it difficult to identify optimal feature subsets within a reasonable timeframe, often resulting in suboptimal or computationally expensive solutions \cite{liu2024large}. Additionally, high-dimensional search spaces exacerbate the problem of the curse of dimensionality, where the distance between data points becomes less informative, making it harder for the algorithm to distinguish between  less and more prominent  feature subsets. This can be more challenging where the feature selection is treated as a sparse optimization problem \cite{sun2024sparse} where most  variables (i.e., feature statuses) are zero. In such cases, every component of optimization process including the initialization, generative operators, and selection can be degraded, necessitating specific tailoring for large-scale scalability \cite{tian2024multi}.

This paper proposes a shrinking scheme to address the challenges of large-scale search spaces by reducing the problem's dimensionality \textit{before the evolutionary process}. As mentioned, large-scale feature selection is inherently a sparse binary optimization problem, where many irrelevant features can be eliminated prior to the expensive evolutionary process. We suggest a ranking-based filtering method that utilizes the correlation of features with class labels and their frequency in a preliminary, inexpensive evolutionary process. This approach aims to identify and remove irrelevant, redundant, and noisy features. The main evolutionary process is then applied to the remaining features, allowing for a more efficient exploration of the search space. 

A serious  issue in large-scale optimization problem is the failure of generative operators, such as crossover and mutation, to consistently produce beneficial variations in the feature subsets \cite{alweshah2021solving}. In high-dimensional spaces, small changes introduced by these operators might not significantly impact the fitness of the feature subsets, leading to slow convergence and less effective feature selection. Moreover, the random nature of these operators can sometimes disrupt useful feature combinations, further complicating the search for optimal solutions\cite{yi2020behavior}. Addressing these challenges is crucial for enhancing the efficiency and effectiveness of evolutionary algorithms in multi-objective feature selection.
To generate more diverse feature subsets, a voting-based crossover scheme has been proposed.  Additionally, for a more intelligent mutation process,  the chance has been given to the features that were prematurely excluded from the population.

In overall, a large-scale multi-objective  method based on the search space shrinking  (termed  LMSSS) is proposed  to address the challenges of sparse  feature selection problems. The major contributions of this study are summarized below:
\begin{itemize}
   \item {Introduction of a shrinking scheme to reduce the dimensionality of the problem before the evolutionary process.}
   \item {Implementation of a ranking-based filtering method to eliminate irrelevant features based on two criteria including their correlation with class labels and their frequency in a lightweight evolutionary process.}
   \item {Development of a smart crossover and mutation scheme tailored to large-scale feature selection problem to enhance the diversity of feature subsets.}
\end{itemize}

\section { Multi-objective Feature Selection}

Multi-objective optimization involves optimizing two or more conflicting objectives simultaneously~\cite{996017}. In the context of feature selection, multi-objective optimization aims to identify a subset of features that balances multiple criteria, such as minimizing the number of selected features while maximizing the classification performance~\cite{bidgoli2021reference}. Dominance is a key concept in multi-objective optimization because it helps compare various solutions based on their performance across all objectives. A solution is said to dominate another if it is at least as good in all objectives and strictly better in at least one objective. If  $\pmb x=(x_{1},x_{2},...,x_{d})$ and  $ \acute{\pmb x}=(\acute{x}_{1},\acute{x}_{2},...,\acute{ x}_{d})$ are two vectors in a minimization problem search space, $\pmb x$ dominates $\acute{\pmb x}$ ($\pmb x\prec\acute{\pmb x}$) if and only if
\begin{eqnarray}
\begin{aligned}
&\forall i\in{\{1,2,...,M\}}, f_i(\pmb x)\leq f_i(\acute{\pmb x}) \wedge\\ 
&\exists j \in{\{1,2,...,M\}}: f_j(\pmb x)<f_j(\acute{\pmb x})
\end{aligned}
\end{eqnarray}

Non-Dominated Sorting (NDS)~\cite{996017} is an algorithm used to rank solutions into different levels based on dominance. Solutions that are not dominated by any other solutions form the first Pareto front. Solutions that are only dominated by those in the first front form the second Pareto front, and so on. NDS helps in identifying and preserving a diverse set of optimal trade-offs between the objectives. This is crucial in multi-objective optimization because it provides decision-makers with a range of choices, each representing a different balance of the objectives. By using NDS, we can identify the best possible solutions that offer a compromise between competing objectives.

In multi-objective feature selection, one common objective is the ratio of selected features, which measures the proportion of features retained from the original feature set. This objective is crucial because a smaller subset of features can lead to simpler models that are easier to interpret and faster to execute. The ratio of features (\( f_1 \)) can be defined as:
\[ f_1 = \frac{\text{Number of selected features}}{\text{Total number of features}} \]
Minimizing \(f_1 \) helps in selecting a smaller number of features, which can reduce the risk of overfitting and improve the generalizability of the model. Therefore, the goal is to minimize \( f_1 \) to achieve a more compact feature set.

Another important objective in feature selection is the quality  of the the feature set for a machine learning task such as classification. Accordingly, F1-score (\( f_2 \)), which is a measure of a machine learning model's accuracy can be consdiered as the second objective.  The F1-score is particularly useful in scenarios with imbalanced datasets, where it is important to balance the trade-off between false positives and false negatives. The F1-score (\( f_2 \)) is defined as:
\[ f_2 = 2 \times \frac{\text{Precision} \times \text{Recall}}{\text{Precision} + \text{Recall}} \]
Maximizing \( f_2 \) ensures that the selected features contribute to a model that accurately identifies the relevant instances while minimizing incorrect classifications. By considering both the ratio of features and the F1-score, multi-objective feature selection seeks to find an optimal subset of features that provides a balance between model simplicity and classification performance.

The overall multi-objective feature selection optimization problem can be expressed as:
\[ \min_{\pmb x} \, \{ f_1(\pmb x), (1-f_2(\pmb x)) \} \],
where $\pmb x$ is a binary vector representing the selected features, \( f_1(\pmb x) \) is the ratio of selected features to the total number of features, and \( f_2(\pmb x) \) is the F1-score. The goal is to find a set of solutions that form the Pareto front, representing the best trade-offs between minimizing the number of selected features and maximizing the F1-score.

\section {Related Works}
To address large-scale feature selection, numerous studies have been conducted in recent years including several review papers \cite{nguyen2020survey, hancer2020survey,dokeroglu2022comprehensive,jiao2023survey}. Some research has focused on sparse optimization problems and applied their methods to large-scale feature selection as a case study. One notable algorithm is SparseEA~\cite{tian2019evolutionary}, which proposes an evolutionary algorithm for large-scale sparse multi-objective optimization problems. This algorithm introduces a novel population initialization strategy by calculating decision variable scores and utilizing genetic operators, considering the sparse nature of Pareto optimal solutions to maintain solution sparsity.

In ~\cite{tian2020pattern}, an evolutionary pattern mining approach is proposed to identify the maximum and minimum candidate sets of nonzero variables in Pareto-optimal solutions. This method limits the dimensions for generating offspring solutions, enhancing performance with a binary crossover and a binary mutation operator to ensure solution sparsity. Similarly, ~\cite{tian2022fast} estimates the sparse distribution of optimal solutions by optimizing a binary vector for each solution and introduces a fast clustering method to significantly reduce the search space dimensionality. SLMEA groups variables based on their level of sparsity (i.e., frequency of appearance in the population). Ding et al.~\cite{ding2022multi} propose a multi-stage knowledge-guided evolutionary algorithm for large-scale sparse multi-objective optimization problems, incorporating diversified sparsity knowledge into the evolutionary process to enhance optimization capability. 

In another category of studies, the primary focus is feature selection as an large-scale optimization problem.  Asilian et al. \cite{bidgoli2022evolutionaryIEEE} proposed a large-scale feature selection framework where the prominence of features is determined by their appearance frequency in several independent runs of an evolutionary process. The most frequent features are then selected to compactly represent a gigapixel histopathology image.
Xu et al.~\cite{xu2020duplication} propose a duplication analysis-based feature selection method, termed DAEA, to minimize duplicated feature subsets. DAEA employs Manhattan distance in the solution space to measure solution similarity. The results demonstrate that the proposed reproduction strategy significantly contributes to DAEA's performance. Cheng et al.~\cite{cheng2021steering} introduce the SMMOEA method, which uses a steering matrix for feature selection, effectively reducing subset size and achieving high accuracy results on 12 high-dimensional datasets.

In the direction of proposing new generative operators for feature selection, Chakraborty \cite{chakraborty2023horizontal} enhanced the Whale Optimization Algorithm (WOA) by incorporating a weight, cooperative learning techniques, and a horizontal crossover strategy into the WOA framework. The introduction of horizontal crossover bolsters the exploration capabilities of WOA, while the integration of cooperative learning techniques and an inertia weight enhances its exploitation abilities. Moreover, a feature-threshold guided crossover operator was introduced in \cite{deng2023feature}. This crossover exchanges features between two parents based on their scores, which are compared based on the error rate among all chromosomes in which the feature appears.
Incorporating filter methods into the evolutionary process is one of the techniques from which many studies have obtained benefits \cite{fu2024mofs, song2021feature, han2023improved}. These methods help in selecting relevant features or individuals based on certain criteria, thereby reducing the search space and improving the efficiency of the optimization process. By filtering out less promising candidates, the evolutionary algorithm can focus on more promising regions of the search space. This approach not only enhances the convergence speed but also increases the likelihood of finding optimal or near-optimal solutions \cite{asghari2023mutual}. Consequently, the integration of filter methods can be instrumental in guiding the optimizer towards more favorable regions, ultimately leading to better performance and results.

\begin{figure*}
\centering

      \includegraphics[width=\linewidth]{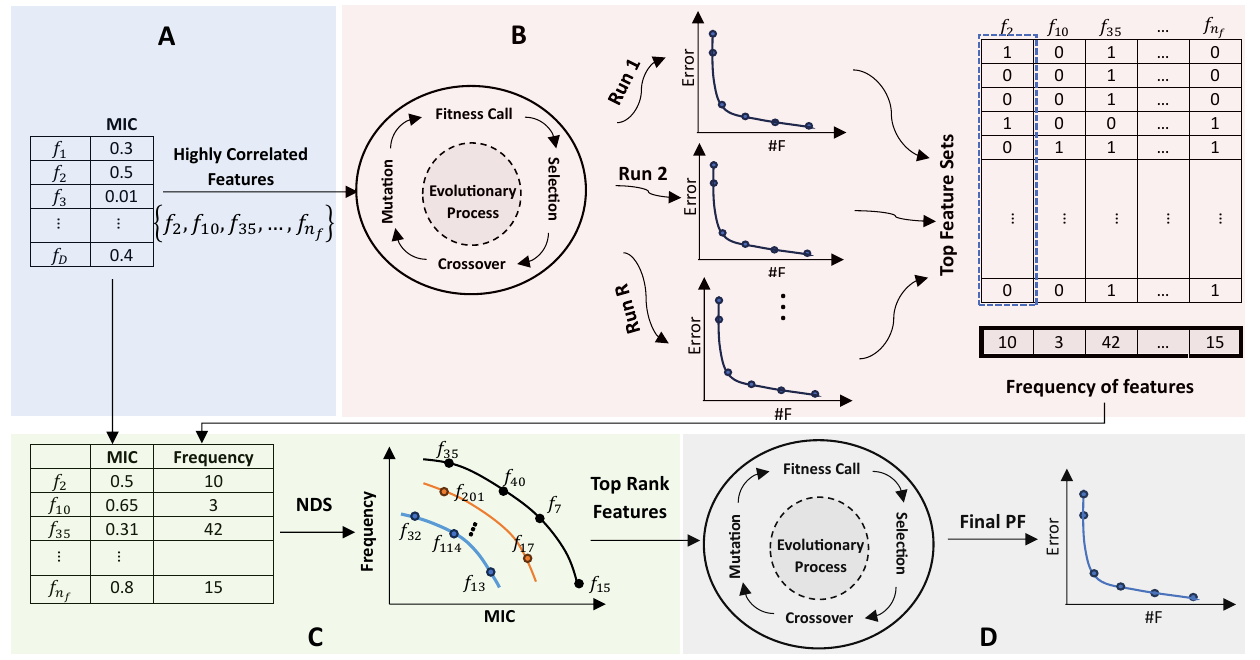} 
     
\caption{Framework overview. The process steps declared with distinct colors include: A. detecting highly correlated features, B. lightweight evolutionary process on $n_f$ top features with highest MIC to select high frequent features from $r$ runs of evolutionary process, C. selecting NDS-based features in terms of frequency and MIC, D. evolutionary process for selecting final set of features on top NDS-based features. }
\label{fig:Framework}
\end{figure*}
\section{Proposed Method}

In this section, the details of the proposed multi-objective framework to select the optimal set of features from a large-scale dataset are described. The algorithm contains two phases: a shrinking process and a multi-objective evolutionary process. In the shrinking process, the search space for the evolutionary algorithm is reduced to alleviate the challenges of large-scale space exploration. 

Multi-objective feature selection is inherently an imbalanced optimization problem in terms of its objectives, as increasing the accuracy of the classification is usually more difficult than reducing the number of features. To tackle this challenge and eliminate irrelevant and redundant features while selecting prominent ones, a correlation-based filter is applied to the entire feature set.  The top-ranked features from this ranking will be moved to a light-weight evolutionary process to extract the frequency of each feature in the output of this process considering the more frequent features seen as more prominent features. Then, as the final step of shrinking phase, NDS applied on 1) correlation values, and 2) the frequency of each feature, as two criteria to rank the filtered  features and consequently the high-ranked features shrink the search space for the second phase.  In the second phase, the main evolutionary process with a novel crossover and mutation is applied to results in a Pareto front as the final output. The steps can be given as bellow:

\begin{itemize}\item \textit{Phase 1: Shrinking Process}
\begin{enumerate}[label=\Alph*.]
\item{Detecting highly Correlated Features}:
    Evaluate and rank each feature based on its correlation with the class label to identify the most relevant features.
    
    \item{Detecting highly Frequent Features}:
    Apply a lightweight evolutionary algorithm to the selected top features from previous step to determine how frequently each feature appears in optimal subsets, indicating their importance.

    \item{NDS-based Feature Ranking}:
  Use non-dominated sorting to rank features based on two criteria: their correlation values and their frequencies from the evolutionary process. This step refines the feature set further by focusing on those that perform well on both criteria.
\end{enumerate}

\item \textit{Phase 2: Multi-Objective Evolutionary Process}
\begin{enumerate}[label=D.]
\item{Main Evolutionary Process}:
 Use the refined set of features from the NDS-based shrinking process as the input for a multi-objective evolutionary algorithm. This algorithm employs novel crossover and mutation strategies to optimize the feature subsets.
\end{enumerate}
\end{itemize}

 An overall view of the algorithm is represented in Figure \ref{fig:Framework}. The details of the algorithm are described in the following subsections.

\subsection{Detecting Highly Correlated  Features}

The first level of elimination filters out features with low correlation to the class label. In large-scale datasets, many features can be irrelevant, and retaining them for the optimization process is considered a waste of the optimize's resources for exploring the search space. Therefore, using a filter criterion like Maximal Information Coefficient (MIC) \cite{reshef2011detecting} can eliminate these features, preventing the evolutionary optimizer from wasting resources on finding optimal solutions. This phase specifically focuses on the primary objective of selecting top-ranked features based on their correlation with the class label.

Mutual Information (MI) \cite{viola1997alignment} is a common method for assessing the relevance between two variables, such as \(X\) and \(Y\). It is defined by the following equation:

\begin{equation}
\text{MI}(X; Y) = \sum_{x \in X} \sum_{y \in Y} p(x, y) \log \frac{p(x, y)}{p(x)p(y)}
\end{equation}

In this equation, \( p(x) \), \( p(y) \), and \( p(x, y) \) represent the probability distributions of \(x\) and \(y\), and their joint distribution, respectively. When \(X\) and \(Y\) are two features, \(\text{MI}(X; Y)\) quantifies the amount of information shared between them. If \(X\) is a feature and \(Y\) represents the class labels, \(\text{MI}(X; Y)\) indicates the degree of dependence between feature \(X\) and the class labels \(Y\).

MIC  is a newly developed statistical measure that uses binning to apply MI to continuous variables. By segmenting the variable values into different grids and normalizing MI, MIC can more accurately measure the relationships between variables.

The calculation of MIC is briefly explained as follows: The original space $G$, comprising the values of variables $X$ and $Y$, is divided into multiple $p$-by-$q$ grids. The characteristic matrix $\text{M}(G)_{p,q}$ represents the highest normalized MI of $G$ within the $p\times q$ partitions, as expressed in the following equation:

\begin{equation}
\text{M}(\text{G})_{p,q} = \frac{\text{max}(\text{MI})}{\text{log} \: \text{min}\{p,q\}}
\end{equation}
max(MI) denotes the maximum MI of G across all $p\times q$ partitions. MIC is then defined as:
\begin{equation}
\text{MIC}= \text{max}_{0 < p\times q <B(n)}\{\text{M(G)}_{p,q}\} \quad \textit{B}(n) = n^{0.6},
\end{equation}

where \(n\) represents the number of instances, and \(\textit{B}(n)\) is the constraint on the grid size $p\times q$. More details and sensitivity analyses of the parameters in MIC can be found in the referenced studies.

The value of MIC tends to approach 1 for two highly correlated variables and approaches 0 for two statistically independent variables. Hence, the higher the MIC value, the greater the dependence between the paired variables.

In order to eliminate irrelevant features, MIC is used to rank all the features in the dataset. The higher the MIC, the more relevant the feature. Higher MIC features typically result in better classification; however, the evolutionary process is responsible for finding the best optimal set. Therefore, $n_{MIC}$ top features with highest MIC values advances to the next step. 

\subsection{Detecting Highly Frequent Features}
The selected features from the previous phase will serve as the input to a lightweight evolutionary process. In fact, the frequent features in several runs of an evolutionary process can be considered prominent features. Therefore, a portion of the budget for the entire feature selection process is allocated to several runs of an evolutionary algorithm to assess the frequency of each feature in the final Pareto front. Despite the stochastic nature of evolutionary algorithms, when a feature is consistently selected in different runs, it can indicate the importance of the corresponding feature.
It is important to note that this algorithm is executed for a limited number of iterations (considered as a lightweight evolutionary process), as this is merely a filtering phase and not the final feature selection phase. The frequency will be used as a metric for ranking the features, and a few iterations are sufficient to compute this frequency. Moreover, in most evolutionary processes, the algorithm usually identifies semi-optimal solutions within just a few iterations after starting. 
Therefore, the evolutionary algorithm described in the next subsection will be applied \( R \) times in \( t \) iterations. Suppose that the size of the population is \( popsize \). Hence, at the end of the process, if we gather all solutions in the final population of each run, \( R \times popsize \) feature subsets will be obtained.

To compute the frequency of each feature, we focus on using the frequency as a metric to rank the features. Since feature subsets with better classification performance are of interest for shrinking space, we consider the top $n_{fs}$\% of solutions from all final populations across different runs to determine the frequency of each feature.  
For each feature, we count its frequency over all these subsets. The frequency value is considered a criterion for feature ranking in the next subsection.
   For each feature \( f \) in the dataset, its frequency \( \text{freq}(f) \) is calculated as:
 \begin{equation}
   \text{freq}(f) = \sum_{i=1}^{n_{fs}\times popsize\times R}  I_{i}(f),
\end{equation}
   where \( I_{i}(f) \) is an indicator function defined as:
   \[
   I_{i}(f) = 
   \begin{cases} 
   1 & \text{if } f \text{ exist in the } i\text{-th subset of the  top sets} \\
   0 & \text{otherwise}
   \end{cases}
   \]

In summary, the process involves running the lightweight evolutionary algorithm 
$R$ times, each for 
$t$ iterations, and then aggregating the results to determine the frequency of each feature, which will be used for ranking in the next step.
\begin{figure*}
\centering
\begin{tabular}{cccc}

      \includegraphics[width=0.3\linewidth]{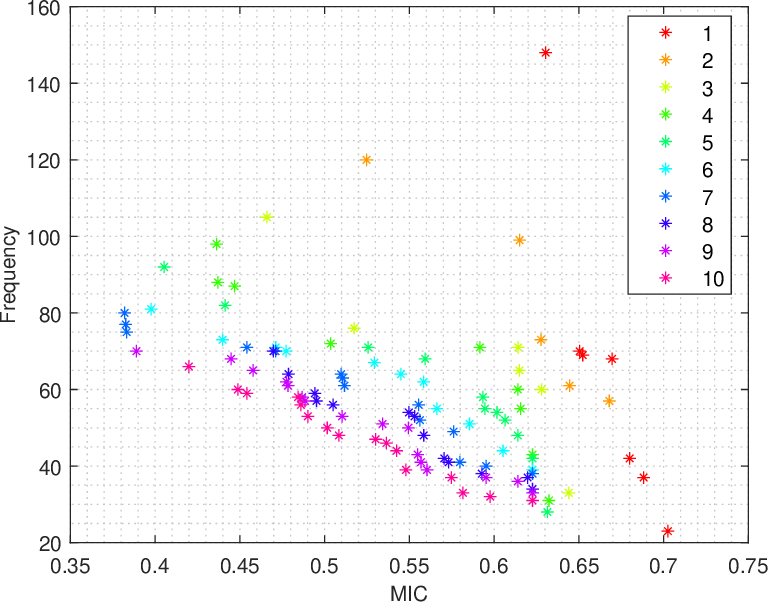} &
     
       \includegraphics[width=0.3\linewidth]{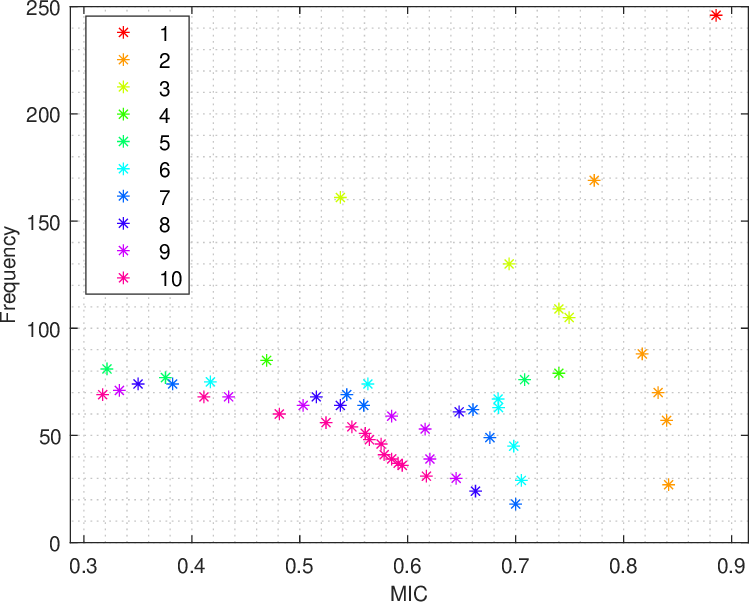} &
    
        \includegraphics[width=0.3\linewidth]{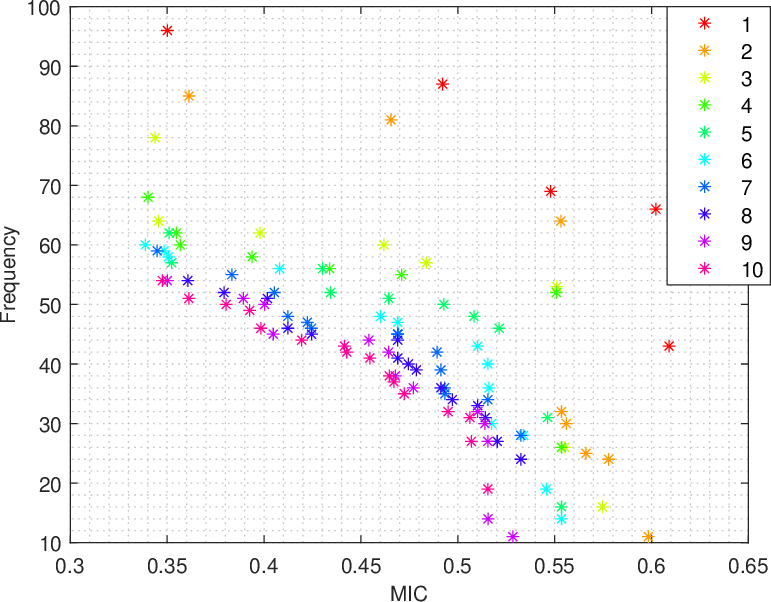} 
               
                       \\
                       
               Carcinom & SRBCT&Lung

\end{tabular}
\caption{NDS-based features. Each plot represents the top features on 10 first ranks based on MIC and frequency using NDS algorithm. The multi-criteria ranking has been visualized in a maximization-maximization Pareto fronts.  }
\label{fig:PFranking}
\end{figure*}
\subsection{NDS-Based Feature Ranking}
All the $f_s$ features from the filter selection phase can be ranked based on two criteria from the two previous steps: MIC  and frequency. The NDS  algorithm is applied to rank these features according to these criteria. Higher ranks correspond to features with high values in both criteria. To select a set of top-ranked features for the evolutionary process, we start from the first Front to choose the desired number of features  (i.e., $n_{NDS}$), creating the shrunk space.

The first criterion, the correlation of the feature with the class label, ensures that the selected feature individually can lead to high classification accuracy. On the other hand, the frequency of the feature in subsets with high classification accuracy represents the impact of the feature in combination with others in the classification task. Therefore, the NDS algorithm evaluates and ranks features based on both individual and collaborative aspects. It is important to note that the frequency value should be scaled to the range $[0,1]$ to match the scale of MIC for NDS ranking. 

In order for a better illustration, Fig. \ref{fig:PFranking} shows the top 10 ranks of the NDS output on some datasets given in the experiments section. As seen, a few features dominate the first ranks, indicating their prominence. In the Carcinom and SRBCT datasets, only one feature in the first rank dominates all other features. This interesting observation reveals that a feature with high correlation can appear in many subsets—150 in Carcinom and 250 in SRBCT. In the Lung dataset, there are four features in the first rank, highlighting the trade-off between frequency and correlation. For instance, a feature on the first front has the maximum frequency (about 100) but a low correlation value (0.36), indicating that even with low correlation, a feature can significantly impact subsets in combination with other features. Conversely, a feature with a correlation value of 0.6 appears in only a few subsets, representing a lower impact in combination with others. The whole process of selecting top ranked features for evolutionary process has been represented in the Algorithm \ref{ranking}.
\begin{algorithm}
\SetAlgoLined
\SetKwInOut{Input}{inputs}\SetKwInOut{Output}{output}
 \Input{ $Dataset$: all features, $D$: number of features, $n_{MIC}$:  number of high correlated  features, $n_{fs}$: number of top feature subsets, $n_{NDS}$: number of top NDS-based features, $R$:  number of runs }
 \Output{ $NS$: selected features}
 \BlankLine
 \tcp{High Correlated Features}
\For{$f\leftarrow 1$ \KwTo $D$}
 {
 $MICScore(f)= $MIC($f$)\;
 }
 $Indices$ = Sort ($MICScore$)\;
 $Dataset$ = $Dataset(Indices(1:n_{MIC}))$\;
  \tcp{High Frequent Features}
 \For{$r\leftarrow 1$ \KwTo $R$}
 {
 $Paretofront$=  Evolutionary Process  ($Dataset$)\;
 $AllParetofronts = AllParetofronts \cup  Paretofront$
 }
 \tcp{sorting Feature Subsets in Terms of Classification Accuracy}
 $Indices$ = Sort ($AllParetofronts$)\;
 $TopSetFeatures$ = $AllParetofronts (Indices(1:n_{fs}))$\;
 \For{$f\leftarrow 1$ \KwTo $D$}
 {
  $FreqScore(f)=\sum_{r=1}^{n_{fs}}TopSetFeatures(f)$\;
 }
 $FeaturesRanking$ = NDS $(MICScore(1:n_{MIC}),FreqScore(1:n_{MIC}))$\;
 $ShrunkSpace$ = $FeaturesRanking(1:n_{NDS}$\;
 \caption{Pseudo-code of shrinking process. }\label{ranking}
\end{algorithm}
\subsection{Evolutionary Process}

In this section, the details of the evolutionary process applied to the top-ranked features from the NDS-based ranking will be discussed. The output of this phase is the final Pareto front of the proposed multi-objective framework. The shrunk space will facilitate a more targeted search process for the optimizer. This not only makes exploration easier but also increases the likelihood of identifying high-quality feature subsets for classification. Moreover, a smaller search space can result in smaller subsets, which further enhances the efficiency of the optimization process.

\subsubsection{Initialization}
Similar to every population-based evolutionary algorithm, the process begins with an initial population containing individuals represented by binary encoding. Each cell of the binary vectors represents the status of a feature, with 1 indicating the presence and 0 indicating the absence of the feature. Therefore, the number of 1's in an individual indicates the number of features in the corresponding individual. Although the impact of initialization is often overlooked in many studies, we will investigate its significant impact, especially in large-scale optimization problems.
To build the initial population, we employ a method to diversify the population in terms of the number of features. In most research, the common method for initialization is to generate a random uniform vector of 0's and 1's, with the probability of each being 50\%. This results in all initial sets having roughly the same size (i.e., 50\% of the total number of features), which can negatively impact the diversity of the population.
Therefore, to provide uniformity at the population level (in addition to the individual level), we generate an integer value $n$ for each individual, representing the size of the feature subset (i.e., the number of selected features). Then, we set $n$ cells to 1 randomly. This method ensures various feature subset sizes across the population, enhancing diversity.

\subsubsection{Generative Operators}
The next important component of the evolutionary process is to generate a set of diverse and high-quality candidate solutions at each generation. One of the challenges in large-scale binary optimization, particularly feature selection, is the degradation of population diversity after a few iterations. Some features are removed in the first iterations and never get a chance to re-enter the population, while others remain in the population, and regular crossover methods, such as uniform crossover \cite{syswerda1989uniform}, simply exchange them between different individuals. This is one of the reasons why redundancy in the population of the feature selection problem is relatively high. Therefore, the effectiveness of the generative operators can play a significant role in advancing the evolutionary process towards promising regions of the search space. Below, the details of the proposed crossover and mutation in order to tackle these challenges are described. 

\textbf{Crossover:} To generate a new offspring from two parents, we propose a novel crossover operator based on a voting mechanism between the two parents. This operator works as follows: two parents, \(P_1\) and \(P_2\), are randomly selected from the population. To prioritize individuals with better classification accuracy, we assume \(P_1\) has better features and a lower classification error (if \(P_2\) is superior, we swap them). The feature selection encoding represents \(P_1\) and \(P_2\) as binary vectors corresponding to the total number of features.

The offspring is generated by aggregating the features of the two parents. Features on which both parents agree are directly transferred to the offspring. Specifically, if both \(P_1\) and \(P_2\) select a feature (value of 1), the offspring will also select that feature. Conversely, if both parents reject a feature (value of 0), the offspring will reject it as well.

\begin{figure}
\centering
\begin{tabular}{cccc}

      \includegraphics[width=0.5\linewidth]{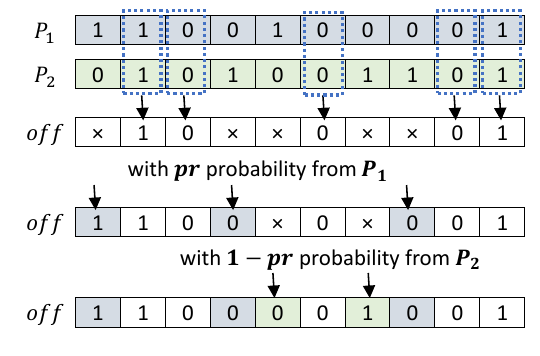}

\end{tabular}
\caption{A sample of crossover. $P_1$ and $P_2$ are two selected parents where $P_1$ is a better subset of feature than $P_2$ in terms of classification accuracy and $pr>0.5$}
\label{fig:Crossover}
\end{figure}

For features where the parents disagree (one parent has a value of 0 and the other has a value of 1), the bit from one of the parents is transferred with probabilities \(Pr\) and \(1-Pr\) for \(P_1\) and \(P_2\), respectively, where \(Pr > 0.5\). This ensures that the offspring is more likely to inherit features from the better-performing parent, thus enhancing the overall quality of the population in terms of classification accuracy. This method is more effective than random uniform bit transferring, especially for the difficult  objective of classification error for feature selection optimization.  
Mathematically, for each feature \(i\), $o_i$:

\[
o_i = \begin{cases} 
1 & \text{if } P_{1i} = 1 \text{ and } P_{2i} = 1, \\
0 & \text{if } P_{1i} = 0 \text{ and } P_{2i} = 0, \\
1 & \text{if } P_{1i} \neq P_{2i} \text{ and } r < Pr, \\
0 & \text{if } P_{1i} \neq P_{2i} \text{ and } r \geq Pr,
\end{cases}
\]
where \(r\) is a random number drawn from a uniform distribution in the range \([0, 1]\).
Fig.~\ref{fig:Crossover} illustrates an example for proposed crossover.

\textbf{Mutation:}
Although, the mutation operator used is the flip bit \cite{chicano2015fitness}, but an additional mutation mechanism is considered to give a chance to features that were excluded in the initial generations and never had the opportunity to return to the population. This issue is particularly prevalent in sparse large-scale binary optimization problems. It has been observed that since one of the objectives in the feature selection problem is to minimize the number of selected features, the optimizer tends to favor smaller feature subsets. Consequently, in a competitive environment, if a feature exits the population due to being part of a combination that results in high classification error, its chance of re-entering the population through mutation is very low. In other words, the population may quickly converge to a small set of features, while some potentially important features are excluded early on.

Since the crossover operator does not introduce new features, mutation is the only operator that can bring new features into the population. To address this issue, we implement an additional mutation mechanism. In the early iterations, if a feature is not selected by any individual in the entire population (i.e., its value is zero across all individuals), we randomly set it to 1 in one of the individuals. This approach ensures that every feature has a chance to be reconsidered, preventing premature convergence to a suboptimal feature subset and enhancing the exploration capability of the algorithm.

\subsubsection{Selection}
The selection phase of the proposed framework is similar to NSGA-II. Therefore, the NDS (Non-dominated Sorting) and crowding distance will be applied to select the top-ranked candidate solutions for the next phase.

It is worth mentioning that the proposed evolutionary process has been employed for both steps: lightweight evolutionary process for selecting the highly frequent features and the final evolutionary process applied to the top-ranked features for the final output.

\section{Experiments}
\subsection{Benchmark Techniques}
The proposed algorithm, LMSSS is compared with the state-of-the-art high-dimensional multi-objective feature selection algorithms.  They include, DAEA~\cite{xu2020duplication}, SparseEA~\cite{tian2019evolutionary}, PMMOEA~\cite{tian2020pattern}, SLMEA~\cite{tian2022fast} SMMOEA~\cite{cheng2021steering}, and MSKEA~\cite{ding2022multi}. Some algorithms such as SparseEA is generally designed for large-scale multi-objective optimization problems but can be employed for binary optimization problems (e.g., feature selection). A brief desciption of each of these algorithms is given in Section III.
All seven algorithms were applied to 15 large-scale datasets represented in Table~\ref{data}.  These 15 datasets exhibit diversity in terms of  classes (ranging from 2 to 16) and instances (ranging from 50 to 203), making them representative of the problem domains addressed by the proposed algorithms. The minimum number of features is 2,000 while they do not exceed 24,481 features. 
\begin{table}[]
\centering
\caption{The information of datasets}
\label{data}
\scriptsize
\begin{tabular}{lllll}
\hline
Number & Dataset       & \#Features & \# Instances & \#Classes \\ \hline
D1      & Colon         & 2,000      & 62           & 2         \\
D2      & SRBCT         & 2,308      & 83           & 4         \\
D3      & lymphoma      & 4,026      & 96           & 9         \\
D4      & Glioma        & 4,434      & 50           & 4         \\
D5      & Leukemia      & 5,147      & 72           & 2         \\
D6      & CNS           & 7,128      & 60           & 2         \\
D7      & DLBCL         & 7,129      & 77           & 2         \\
D8      & Leukemia\_all & 7,129      & 72           & 2         \\
D9      & Carcinom      & 9,182      & 174          & 11        \\
D10     & Prostate      & 10,509     & 102          & 2         \\
D11     & CLL\_SUB      & 11,340     & 111          & 3         \\
D12     & 11Tumor       & 12,533     & 174          & 11        \\
D13     & Lung          & 12,600     & 203          & 5         \\
D14     & GLI           & 22,283     & 85           & 2         \\
D15     & Breast        & 24,481     & 97           & 2         \\ \hline
\end{tabular}
\end{table}

To compare the performance of different EMO-based feature selection methods, two indicators, namely, the hypervolume (HV) \cite{shang2020survey} and the inverted generational distance (IGD) \cite{sun2018igd}, were utilized. Following each run, each algorithm obtained a Pareto front on training set. The feature subsets  on training  Pareto front are evaluated on test set resulting in test Pareto front. For each run,  the values of performance indicators are calculated on resultant test Pareto front. The average of indicators over 31 independent runs are reported in results section. In order to calculate the IGD, we gather and merge all the output feature subsets generated by the seven algorithms into a single set. Subsequently, all the feature subsets in this combined set are ranked using NDS algorithm. The first PF is retained as the optimal Pareto front for IGD. Higher HV values or smaller IGD values indicate superior feature subsets. 
In addition to IGD and HV, the minimum value of classification error (MCE) on test Pareto front  is also represented. 
To assess the significance of LMSSS relative to the other algorithms, the Wilcoxon Rank-sum test \cite{de2018wilcoxon} with a significance level of 0.05 was employed. On each table, the signs $\uparrow$, $\downarrow$, and $^\circ$ indicate that the LMSSS  is significantly better than, worse than, or has no significant difference
from the competitor algorithm, respectively. The bold value indicates the highest number achievable by an algorithm.

\subsection{Parameter Settings}
 For each dataset, every algorithm was executed independently 31 times, with the data randomly divided into a training set (70\%) and a test set (30\%). In order to provide a fair comparison, the same training and test sets are considered for all algorithms.  During the optimization process, $k$-nearest neighbors ($k$-NN) \cite{kramer2013k} with one-leave-out cross validation on the training set is employed to compute the classification error rate. The value of $k$ in $k$-NN is set to five to strike a balance between accuracy and efficiency. The total number of generations for all algorithms is set to 100 times while the population size is set to 200. For LMSSS, 50\% of the budget is allocated to a lightweight evolutionary process, which is independently repeated 5 times with 10\% of the budget (i.e., 10 iterations) for each run. The remaining 50\% of the budget is allocated to the main evolutionary process on the shrunk space to generate the final results. The parameters $n_{MIC}$ and $n_{NDS}$ are set to 1000 and 200, respectively. These values were selected based on our experiments, where smaller values for even the smallest datasets were too much filtering and  larger values for larger datasets did not result in better outputs. Thus, we maintained constant values for the number of features in the shrunk space. The parameter $n_{fs}$ is set to 50\% of the total number of solutions on all final soltuions from the 5 runs of the lightweight evolutionary process.

The parameter $n_m$ is set to 10\% of the total number of iterations, and the mutation rate is set to $1/D$, where $D$ is the total number of features in a dataset. The parameter $pr$ is set to 0.7. The crossover rate for regular NSGA-II is set to 0.5. For each competitor algorithm, parameter values are taken from their respective reference papers.

\subsection {Results}

\subsubsection {Analysis on Multi-objective Evaluation Metrics }
Table~\ref{tab:HV} displays the HV values for 7 benchmark algorithms on the test set of 15 large-scale datasets. In the final row of the table, you can see the overall rank for each algorithm. Notably, LMSSS achieves the highest rank among all benchmark algorithms, while MSKEA takes the last position. DAEA secures the second rank, with 6 losses and 9 ties compared to LMSSS. It's worth noting that LMSSS consistently outperforms other algorithms, never losing in comparison. In other words, its HV is either higher or equal to other algorithms across all benchmarks indicating more spread  and distribution over the Pareto front. In summary, LMSSS wins 63 and draws 27 with no lose out of the 90 comparisons for the test HV results.

On the other hand, Table~\ref{tab:IGD} shows the IGD values for all algorithms. Again, LMSSS obtained the first rank in overall with a large margin from the others while its only significant lose  is against DAEA on SRBCT dataset. In overall, the minimum number of wins of LMSSS is 9 against SparseEA while the DAEA is placed in the second position. In summary, LMSSS wins 70, draws 19, and loses 1 out of the 90 comparisons for the test IGD results.
\begin{table*}[]
\centering
\setlength{\tabcolsep}{2.5pt}
\caption{ HV values (the higher, the better) achieved by the competitor algorithms. The highest value in each dataset is highlighted in bold. The symbols $\uparrow$, $\downarrow$, and $^\circ$ indicate that LMSSS is significantly better, significantly worse, or not significantly different from the competitor algorithm, respectively. The last row displays the rank of each algorithm, with lower ranks indicating better performance, as determined by the Friedman test. Additionally, the number of wins/ties/losses of LMSSS compared to the other algorithms is shown in the final row. }
\label{tab:HV}
\scriptsize
\begin{tabular}{llllllll}
\hline
       & DAEA                                                             & SparseEA                                                         & PMMOEA                                                           & SLMEA                                                            & SMMOEA                                                           & MSKEA                                                            & LMSSS                                                             \\ \hline
D1         & \begin{tabular}[c]{@{}l@{}}7.977E-01   $^\circ$\\ ±7.446E-02\end{tabular} & \begin{tabular}[c]{@{}l@{}}6.890E-01  $\downarrow$ \\ ±6.659E-02\end{tabular} & \begin{tabular}[c]{@{}l@{}}7.575E-01  $\downarrow$ \\ ±8.907E-02\end{tabular} & \begin{tabular}[c]{@{}l@{}}7.364E-01 $\downarrow$ \\  ±8.402E-02\end{tabular} & \begin{tabular}[c]{@{}l@{}}7.416E-01  $\downarrow$ \\ ±7.609E-02\end{tabular} & \begin{tabular}[c]{@{}l@{}}6.715E-01 $\downarrow$ \\  ±4.291E-02\end{tabular} & \begin{tabular}[c]{@{}l@{}}\textbf{8.305E-01} \\  ±3.491E-02\end{tabular} \\ \hline
D2         & \begin{tabular}[c]{@{}l@{}}9.154E-01  $^\circ$\\ ±3.059E-02\end{tabular} & \begin{tabular}[c]{@{}l@{}}8.856E-01 $\downarrow$ \\  ±4.926E-02\end{tabular} & \begin{tabular}[c]{@{}l@{}}9.154E-01  $^\circ$\\ ±4.227E-02\end{tabular} & \begin{tabular}[c]{@{}l@{}}8.967E-01  $\downarrow$ \\ ±3.810E-02\end{tabular} & \begin{tabular}[c]{@{}l@{}}9.041E-01 $\downarrow$ \\  ±5.016E-02\end{tabular} & \begin{tabular}[c]{@{}l@{}}8.630E-01  $\downarrow$ \\ ±3.723E-02\end{tabular} & \begin{tabular}[c]{@{}l@{}}\textbf{9.370E-01}  \\ ±2.956E-02\end{tabular} \\ \hline
D3      & \begin{tabular}[c]{@{}l@{}}7.538E-01 $\downarrow$ \\  ±5.358E-02\end{tabular} & \begin{tabular}[c]{@{}l@{}}7.495E-01  $\downarrow$ \\ ±4.004E-02\end{tabular} & \begin{tabular}[c]{@{}l@{}}7.652E-01 $^\circ$\\  ±5.664E-02\end{tabular} & \begin{tabular}[c]{@{}l@{}}7.454E-01 $\downarrow$ \\  ±6.116E-02\end{tabular} & \begin{tabular}[c]{@{}l@{}}7.693E-01  $^\circ$\\ ±4.449E-02\end{tabular} & \begin{tabular}[c]{@{}l@{}}7.679E-01  $^\circ$\\ ±5.239E-02\end{tabular} & \begin{tabular}[c]{@{}l@{}}\textbf{7.921E-01}  \\ ±3.794E-02\end{tabular} \\ \hline
D4        & \begin{tabular}[c]{@{}l@{}}6.924E-01 $\downarrow$ \\  ±8.468E-02\end{tabular} & \begin{tabular}[c]{@{}l@{}}6.830E-01 $\downarrow$ \\  ±9.240E-02\end{tabular} & \begin{tabular}[c]{@{}l@{}}7.257E-01 $^\circ$\\  ±8.112E-02\end{tabular} & \begin{tabular}[c]{@{}l@{}}6.849E-01  $\downarrow$ \\ ±9.377E-02\end{tabular} & \begin{tabular}[c]{@{}l@{}}6.664E-01 $\downarrow$\\   ±8.748E-02\end{tabular} & \begin{tabular}[c]{@{}l@{}}6.257E-01 $\downarrow$ \\  ±7.133E-02\end{tabular} & \begin{tabular}[c]{@{}l@{}}\textbf{7.551E-01}  \\  ±4.272E-02\end{tabular} \\ \hline
D5  & \begin{tabular}[c]{@{}l@{}}9.273E-01  $^\circ$\\ ±4.613E-02\end{tabular} & \begin{tabular}[c]{@{}l@{}}9.012E-01 $\downarrow$ \\  ±4.410E-02\end{tabular} & \begin{tabular}[c]{@{}l@{}}9.375E-01 $^\circ$\\  ±4.934E-02\end{tabular} & \begin{tabular}[c]{@{}l@{}}9.128E-01$^\circ$\\   ±5.821E-02\end{tabular} & \begin{tabular}[c]{@{}l@{}}8.911E-01  $\downarrow$ \\ ±4.672E-02\end{tabular} & \begin{tabular}[c]{@{}l@{}}9.128E-01  $^\circ$\\ ±3.785E-02\end{tabular} & \begin{tabular}[c]{@{}l@{}}\textbf{9.446E-01} \\  ±4.258E-02\end{tabular} \\ \hline
D6           & \begin{tabular}[c]{@{}l@{}}7.156E-01 $^\circ$\\  ±9.731E-02\end{tabular} & \begin{tabular}[c]{@{}l@{}}6.647E-01 $\downarrow$ \\  ±8.341E-02\end{tabular} & \begin{tabular}[c]{@{}l@{}}6.876E-01 $^\circ$\\  ±7.808E-02\end{tabular} & \begin{tabular}[c]{@{}l@{}}6.841E-01  $\downarrow$ \\ ±7.816E-02\end{tabular} & \begin{tabular}[c]{@{}l@{}}6.788E-01 $\downarrow$ \\  ±7.979E-02\end{tabular} & \begin{tabular}[c]{@{}l@{}}5.841E-01  $\downarrow$ \\ ±2.117E-02\end{tabular} & \begin{tabular}[c]{@{}l@{}}\textbf{7.348E-01}  \\ ±6.249E-02\end{tabular} \\ \hline
D7         & \begin{tabular}[c]{@{}l@{}}9.109E-01  $^\circ$\\ ±5.329E-02\end{tabular} & \begin{tabular}[c]{@{}l@{}}9.136E-01 $^\circ$\\  ±4.639E-02\end{tabular} & \begin{tabular}[c]{@{}l@{}}9.151E-01$^\circ$\\   ±6.138E-02\end{tabular} & \begin{tabular}[c]{@{}l@{}}9.040E-01 $^\circ$\\  ±6.392E-02\end{tabular} & \begin{tabular}[c]{@{}l@{}}8.929E-01 $^\circ$\\  ±6.537E-02\end{tabular} & \begin{tabular}[c]{@{}l@{}}8.608E-01  $\downarrow$ \\ ±4.414E-02\end{tabular} & \begin{tabular}[c]{@{}l@{}}\textbf{9.233E-01}  \\ ±3.645E-02\end{tabular} \\ \hline
D8 & \begin{tabular}[c]{@{}l@{}}9.317E-01 $^\circ$\\  ±5.437E-02\end{tabular} & \begin{tabular}[c]{@{}l@{}}8.897E-01  $\downarrow$ \\ ±3.561E-02\end{tabular} & \begin{tabular}[c]{@{}l@{}}9.158E-01 $\downarrow$ \\   ±5.585E-02\end{tabular} & \begin{tabular}[c]{@{}l@{}}9.143E-01  $\downarrow$ \\ ±4.183E-02\end{tabular} & \begin{tabular}[c]{@{}l@{}}9.013E-01  $\downarrow$ \\ ±5.226E-02\end{tabular} & \begin{tabular}[c]{@{}l@{}}8.941E-01 $\downarrow$ \\  ±2.951E-02\end{tabular} & \begin{tabular}[c]{@{}l@{}}\textbf{9.649E-01} \\  ±3.680E-02\end{tabular} \\ \hline
D9      & \begin{tabular}[c]{@{}l@{}}8.358E-01  $\downarrow$ \\ ±3.692E-02\end{tabular} & \begin{tabular}[c]{@{}l@{}}7.915E-01  $\downarrow$ \\ ±3.850E-02\end{tabular} & \begin{tabular}[c]{@{}l@{}}7.887E-01 $\downarrow$ \\  ±4.442E-02\end{tabular} & \begin{tabular}[c]{@{}l@{}}7.939E-01  $\downarrow$ \\ ±3.670E-02\end{tabular} & \begin{tabular}[c]{@{}l@{}}7.822E-01 $\downarrow$ \\  ±3.432E-02\end{tabular} & \begin{tabular}[c]{@{}l@{}}7.526E-01 $\downarrow$ \\  ±3.818E-02\end{tabular} & \begin{tabular}[c]{@{}l@{}}\textbf{8.708E-01}\\  ±2.057E-02\end{tabular}  \\ \hline
D10      & \begin{tabular}[c]{@{}l@{}}8.279E-01 $\downarrow$ \\  ±3.912E-02\end{tabular} & \begin{tabular}[c]{@{}l@{}}8.139E-01 $\downarrow$ \\  ±3.661E-02\end{tabular} & \begin{tabular}[c]{@{}l@{}}8.451E-01 $^\circ$\\   ±3.918E-02\end{tabular} & \begin{tabular}[c]{@{}l@{}}7.999E-01  $\downarrow$ \\ ±3.918E-02\end{tabular} & \begin{tabular}[c]{@{}l@{}}8.149E-01  $\downarrow$ \\ ±4.313E-02\end{tabular} & \begin{tabular}[c]{@{}l@{}}8.010E-01  $\downarrow$ \\ ±3.501E-02\end{tabular} & \begin{tabular}[c]{@{}l@{}}\textbf{8.686E-01} \\ ±3.272E-02\end{tabular}  \\ \hline
D11      & \begin{tabular}[c]{@{}l@{}}7.218E-01 $^\circ$\\  ±6.136E-02\end{tabular} & \begin{tabular}[c]{@{}l@{}}6.704E-01 $\downarrow$ \\  ±3.879E-02\end{tabular} & \begin{tabular}[c]{@{}l@{}}6.799E-01  $\downarrow$ \\ ±6.204E-02\end{tabular} & \begin{tabular}[c]{@{}l@{}}6.666E-01  $\downarrow$ \\ ±6.155E-02\end{tabular} & \begin{tabular}[c]{@{}l@{}}6.513E-01  $\downarrow$ \\ ±7.911E-02\end{tabular} & \begin{tabular}[c]{@{}l@{}}6.514E-01 $\downarrow$ \\  ±3.629E-02\end{tabular} & \begin{tabular}[c]{@{}l@{}}\textbf{7.551E-01} \\  ±5.179E-02\end{tabular} \\ \hline
D12       & \begin{tabular}[c]{@{}l@{}}7.899E-01  $\downarrow$ \\ ±4.267E-02\end{tabular} & \begin{tabular}[c]{@{}l@{}}7.468E-01 $\downarrow$ \\  ±5.238E-02\end{tabular} & \begin{tabular}[c]{@{}l@{}}7.503E-01 $\downarrow$ \\  ±5.127E-02\end{tabular} & \begin{tabular}[c]{@{}l@{}}7.451E-01  $\downarrow$ \\ ±4.400E-02\end{tabular} & \begin{tabular}[c]{@{}l@{}}7.395E-01 $\downarrow$ \\  ±3.991E-02\end{tabular} & \begin{tabular}[c]{@{}l@{}}7.144E-01 $\downarrow$ \\  ±5.533E-02\end{tabular} & \begin{tabular}[c]{@{}l@{}}\textbf{8.513E-01} \\  ±2.616E-02\end{tabular} \\ \hline
D13          & \begin{tabular}[c]{@{}l@{}}9.052E-01  $\downarrow$ \\ ±2.411E-02\end{tabular} & \begin{tabular}[c]{@{}l@{}}8.787E-01  $\downarrow$ \\ ±2.551E-02\end{tabular} & \begin{tabular}[c]{@{}l@{}}8.967E-01  $\downarrow$ \\ ±2.456E-02\end{tabular} & \begin{tabular}[c]{@{}l@{}}8.851E-01  $\downarrow$ \\ ±3.219E-02\end{tabular} & \begin{tabular}[c]{@{}l@{}}8.876E-01 $\downarrow$ \\  ±3.172E-02\end{tabular} & \begin{tabular}[c]{@{}l@{}}8.629E-01 $\downarrow$ \\  ±3.294E-02\end{tabular} & \begin{tabular}[c]{@{}l@{}}\textbf{9.384E-01} \\  ±2.758E-02\end{tabular} \\ \hline
D14           & \begin{tabular}[c]{@{}l@{}}\textbf{8.846E-01} $^\circ$\\  ±4.402E-02\end{tabular} & \begin{tabular}[c]{@{}l@{}}7.320E-01 $\downarrow$ \\  ±6.933E-03\end{tabular} & \begin{tabular}[c]{@{}l@{}}8.756E-01  $^\circ$\\ ±3.736E-02\end{tabular} & \begin{tabular}[c]{@{}l@{}}8.512E-01  $^\circ$\\ ±5.507E-02\end{tabular} & \begin{tabular}[c]{@{}l@{}}8.601E-01 $^\circ$\\  ±5.198E-02\end{tabular} & \begin{tabular}[c]{@{}l@{}}7.357E-01  $\downarrow$ \\ ±2.733E-02\end{tabular} & \begin{tabular}[c]{@{}l@{}}8.743E-01  \\ ±3.903E-02\end{tabular} \\ \hline
D15        & \begin{tabular}[c]{@{}l@{}}7.044E-01 $^\circ$\\  ±4.853E-02\end{tabular} & \begin{tabular}[c]{@{}l@{}}6.000E-01  $\downarrow$ \\ ±4.517E-16\end{tabular} & \begin{tabular}[c]{@{}l@{}}6.900E-01 $^\circ$\\  ±6.956E-02\end{tabular} & \begin{tabular}[c]{@{}l@{}}6.711E-01 $\downarrow$ \\  ±7.097E-02\end{tabular} & \begin{tabular}[c]{@{}l@{}}6.532E-01 $\downarrow$ \\  ±5.714E-02\end{tabular} & \begin{tabular}[c]{@{}l@{}}6.000E-01  $\downarrow$ \\ ±4.517E-16\end{tabular} & \begin{tabular}[c]{@{}l@{}}\textbf{7.244E-01} \\  ±5.099E-02\end{tabular} \\ \hline
R.       & 3.62 (6/9/0)                                                     & 5.28 (14/1/0)                                                    & 3.98 (6/9/0)                                                     & 4.68 (12/3/0)                                                    & 4.90 (12/3/0)                                                    & 5.87 (13/2/0)                                                    & \textbf{2.970}                                                            \\ \hline
\end{tabular}
\end{table*}

\begin{table*}[]
\centering

\caption{IGD values (the higher, the better) achieved by the competitor algorithms. The highest value in each dataset is highlighted in bold. The symbols $\uparrow$, $\downarrow$, and $^\circ$ indicate that LMSSS is significantly better, significantly worse, or not significantly different from the competitor algorithm, respectively. The last row displays the rank of each algorithm, with lower ranks indicating better performance, as determined by the Friedman test. Additionally, the number of wins/ties/losses of LMSSS compared to the other algorithms is shown in the final row.}
\label{tab:IGD}
\scriptsize
\setlength{\tabcolsep}{2.4pt}
\begin{tabular}{llllllll}
\hline
       & DAEA                                                             & SparseEA                                                         & PMMOEA                                                           & SLMEA                                                            & SMMOEA                                                           & MSKEA                                                            & LMSSS                                                             \\ \hline
D1         & \begin{tabular}[c]{@{}l@{}}7.690E-02  $\downarrow$\\ ±4.465E-02\end{tabular} & \begin{tabular}[c]{@{}l@{}}1.535E-01  $\downarrow$\\ ±5.784E-02\end{tabular} & \begin{tabular}[c]{@{}l@{}}1.016E-01  $\downarrow$\\ ±6.701E-02\end{tabular} & \begin{tabular}[c]{@{}l@{}}1.165E-01  $\downarrow$\\ ±6.489E-02\end{tabular} & \begin{tabular}[c]{@{}l@{}}1.112E-01  $\downarrow$\\ ±5.507E-02\end{tabular} & \begin{tabular}[c]{@{}l@{}}1.669E-01  $\downarrow$\\ ±4.194E-02\end{tabular} & \begin{tabular}[c]{@{}l@{}}\textbf{4.422E-02}  \\ ±1.681E-02\end{tabular} \\ \hline
D2         & \begin{tabular}[c]{@{}l@{}}\textbf{3.564E-02}  $\uparrow$\\\ ±1.382E-02\end{tabular} & \begin{tabular}[c]{@{}l@{}}5.198E-02  $^\circ$\\ ±2.003E-02\end{tabular} & \begin{tabular}[c]{@{}l@{}}5.047E-02  $^\circ$\\ ±2.582E-02\end{tabular} & \begin{tabular}[c]{@{}l@{}}5.543E-02  $^\circ$\\ ±2.035E-02\end{tabular} & \begin{tabular}[c]{@{}l@{}}5.984E-02  $^\circ$\\ ±2.130E-02\end{tabular} & \begin{tabular}[c]{@{}l@{}}6.274E-02  $\downarrow$\\ ±1.502E-02\end{tabular} & \begin{tabular}[c]{@{}l@{}}5.245E-02  \\ ±2.227E-02\end{tabular} \\ \hline
D3      & \begin{tabular}[c]{@{}l@{}}1.215E-01 $\downarrow$\\  ±4.268E-02\end{tabular} & \begin{tabular}[c]{@{}l@{}}1.225E-01  $\downarrow$\\ ±2.977E-02\end{tabular} & \begin{tabular}[c]{@{}l@{}}1.153E-01  $^\circ$\\ ±3.988E-02\end{tabular} & \begin{tabular}[c]{@{}l@{}}1.294E-01  $\downarrow$\\ ±4.834E-02\end{tabular} & \begin{tabular}[c]{@{}l@{}}1.096E-01  $^\circ$\\ ±2.960E-02\end{tabular} & \begin{tabular}[c]{@{}l@{}}1.146E-01  $^\circ$\\ ±3.521E-02\end{tabular} & \begin{tabular}[c]{@{}l@{}}\textbf{9.839E-02}  \\ ±1.959E-02\end{tabular} \\ \hline
D4        & \begin{tabular}[c]{@{}l@{}}1.767E-01  $\downarrow$\\ ±7.773E-02\end{tabular} & \begin{tabular}[c]{@{}l@{}}1.901E-01  $\downarrow$\\ ±7.841E-02\end{tabular} & \begin{tabular}[c]{@{}l@{}}1.526E-01  $^\circ$\\ ±6.253E-02\end{tabular} & \begin{tabular}[c]{@{}l@{}}1.871E-01  $\downarrow$\\ ±8.236E-02\end{tabular} & \begin{tabular}[c]{@{}l@{}}2.008E-01  $\downarrow$\\ ±8.243E-02\end{tabular} & \begin{tabular}[c]{@{}l@{}}2.389E-01 $\downarrow$\\  ±6.978E-02\end{tabular} & \begin{tabular}[c]{@{}l@{}}\textbf{1.178E-01} \\ ±3.596E-02\end{tabular}  \\ \hline
D5  & \begin{tabular}[c]{@{}l@{}}8.116E-02  $\downarrow$\\ ±2.979E-02\end{tabular} & \begin{tabular}[c]{@{}l@{}}6.366E-02  $\downarrow$\\ ±1.778E-02\end{tabular} & \begin{tabular}[c]{@{}l@{}}7.680E-02  $\downarrow$\\ ±3.183E-02\end{tabular} & \begin{tabular}[c]{@{}l@{}}8.278E-02  $\downarrow$\\ ±3.113E-02\end{tabular} & \begin{tabular}[c]{@{}l@{}}7.544E-02  $\downarrow$\\ ±2.523E-02\end{tabular} & \begin{tabular}[c]{@{}l@{}}5.916E-02  $^\circ$\\ ±1.258E-02\end{tabular} & \begin{tabular}[c]{@{}l@{}}\textbf{5.302E-02}  \\ ±3.151E-02\end{tabular} \\ \hline
D6           & \begin{tabular}[c]{@{}l@{}}1.322E-01  $^\circ$\\ ±8.380E-02\end{tabular} & \begin{tabular}[c]{@{}l@{}}1.758E-01  $\downarrow$\\ ±8.002E-02\end{tabular} & \begin{tabular}[c]{@{}l@{}}1.534E-01  $\downarrow$\\ ±7.459E-02\end{tabular} & \begin{tabular}[c]{@{}l@{}}1.582E-01  $\downarrow$\\ ±7.114E-02\end{tabular} & \begin{tabular}[c]{@{}l@{}}1.627E-01  $\downarrow$\\ ±7.461E-02\end{tabular} & \begin{tabular}[c]{@{}l@{}}2.543E-01  $\downarrow$\\ ±2.119E-02\end{tabular} & \begin{tabular}[c]{@{}l@{}}\textbf{1.111E-01}  \\ ±5.037E-02\end{tabular} \\ \hline
D7         & \begin{tabular}[c]{@{}l@{}}7.765E-02  $\downarrow$\\ ±2.883E-02\end{tabular} & \begin{tabular}[c]{@{}l@{}}6.834E-02  $^\circ$\\ ±2.668E-02\end{tabular} & \begin{tabular}[c]{@{}l@{}}7.179E-02  $^\circ$\\ ±3.476E-02\end{tabular} & \begin{tabular}[c]{@{}l@{}}7.005E-02  $^\circ$\\ ±3.820E-02\end{tabular} & \begin{tabular}[c]{@{}l@{}}8.141E-02  $\downarrow$\\ ±3.664E-02\end{tabular} & \begin{tabular}[c]{@{}l@{}}9.777E-02  $\downarrow$\\ ±2.515E-02\end{tabular} & \begin{tabular}[c]{@{}l@{}}\textbf{6.220E-02}  \\ ±2.298E-02\end{tabular} \\ \hline
D8 & \begin{tabular}[c]{@{}l@{}}7.666E-02  $\downarrow$\\ ±1.372E-02\end{tabular} & \begin{tabular}[c]{@{}l@{}}8.947E-02  $^\circ$\\ ±1.174E-02\end{tabular} & \begin{tabular}[c]{@{}l@{}}8.234E-02  $\downarrow$\\ ±1.832E-02\end{tabular} & \begin{tabular}[c]{@{}l@{}}7.666E-02  $\downarrow$\\ ±2.453E-02\end{tabular} & \begin{tabular}[c]{@{}l@{}}7.854E-02  $\downarrow$\\ ±2.266E-02\end{tabular} & \begin{tabular}[c]{@{}l@{}}8.621E-02  $\downarrow$\\ ±1.159E-02\end{tabular} & \begin{tabular}[c]{@{}l@{}}\textbf{5.372E-02}  \\ ±2.107E-02\end{tabular} \\ \hline
D9      & \begin{tabular}[c]{@{}l@{}}3.256E-02  $\downarrow$\\ ±1.843E-02\end{tabular} & \begin{tabular}[c]{@{}l@{}}5.879E-02  $\downarrow$\\ ±2.531E-02\end{tabular} & \begin{tabular}[c]{@{}l@{}}6.095E-02  $\downarrow$\\ ±2.605E-02\end{tabular} & \begin{tabular}[c]{@{}l@{}}5.581E-02  $\downarrow$\\ ±2.558E-02\end{tabular} & \begin{tabular}[c]{@{}l@{}}6.269E-02  $\downarrow$\\ ±2.513E-02\end{tabular} & \begin{tabular}[c]{@{}l@{}}8.716E-02  $\downarrow$\\ ±3.251E-02\end{tabular} & \begin{tabular}[c]{@{}l@{}}\textbf{2.091E-02}  \\ ±7.557E-03\end{tabular} \\ \hline
D10      & \begin{tabular}[c]{@{}l@{}}7.786E-02  $\downarrow$\\ ±1.842E-02\end{tabular} & \begin{tabular}[c]{@{}l@{}}8.471E-02  $\downarrow$\\ ±1.803E-02\end{tabular} & \begin{tabular}[c]{@{}l@{}}7.284E-02  $^\circ$\\ ±2.006E-02\end{tabular} & \begin{tabular}[c]{@{}l@{}}9.041E-02  $\downarrow$\\ ±2.058E-02\end{tabular} & \begin{tabular}[c]{@{}l@{}}8.643E-02  $\downarrow$\\ ±2.719E-02\end{tabular} & \begin{tabular}[c]{@{}l@{}}9.002E-02  $\downarrow$\\ ±1.700E-02\end{tabular} & \begin{tabular}[c]{@{}l@{}}\textbf{6.281E-02}  \\ ±2.393E-02\end{tabular} \\ \hline
D11      & \begin{tabular}[c]{@{}l@{}}1.166E-01  $^\circ$\\ ±5.595E-02\end{tabular} & \begin{tabular}[c]{@{}l@{}}1.639E-01  $\downarrow$\\ ±3.880E-02\end{tabular} & \begin{tabular}[c]{@{}l@{}}1.554E-01  $\downarrow$\\ ±6.017E-02\end{tabular} & \begin{tabular}[c]{@{}l@{}}1.683E-01  $\downarrow$\\ ±6.022E-02\end{tabular} & \begin{tabular}[c]{@{}l@{}}1.843E-01  $\downarrow$\\ ±7.637E-02\end{tabular} & \begin{tabular}[c]{@{}l@{}}1.829E-01  $\downarrow$\\ ±3.630E-02\end{tabular} & \begin{tabular}[c]{@{}l@{}}\textbf{9.043E-02}  \\ ±3.311E-02\end{tabular} \\ \hline
D12       & \begin{tabular}[c]{@{}l@{}}4.958E-02  $\downarrow$\\ ±2.189E-02\end{tabular} & \begin{tabular}[c]{@{}l@{}}7.909E-02  $\downarrow$\\ ±3.130E-02\end{tabular} & \begin{tabular}[c]{@{}l@{}}7.002E-02  $\downarrow$\\ ±3.373E-02\end{tabular} & \begin{tabular}[c]{@{}l@{}}7.406E-02  $\downarrow$\\ ±2.922E-02\end{tabular} & \begin{tabular}[c]{@{}l@{}}8.742E-02  $\downarrow$\\ ±2.405E-02\end{tabular} & \begin{tabular}[c]{@{}l@{}}9.709E-02  $\downarrow$\\ ±3.660E-02\end{tabular} & \begin{tabular}[c]{@{}l@{}}\textbf{2.553E-02}  \\ ±6.114E-03\end{tabular} \\ \hline
D13          & \begin{tabular}[c]{@{}l@{}}4.349E-02  $\downarrow$\\ ±1.254E-02\end{tabular} & \begin{tabular}[c]{@{}l@{}}5.675E-02  $\downarrow$\\ ±1.373E-02\end{tabular} & \begin{tabular}[c]{@{}l@{}}4.340E-02  $\downarrow$\\ ±1.379E-02\end{tabular} & \begin{tabular}[c]{@{}l@{}}5.147E-02  $\downarrow$\\ ±1.476E-02\end{tabular} & \begin{tabular}[c]{@{}l@{}}5.791E-02  $\downarrow$\\ ±1.842E-02\end{tabular} & \begin{tabular}[c]{@{}l@{}}6.882E-02  $\downarrow$\\ ±1.557E-02\end{tabular} & \begin{tabular}[c]{@{}l@{}}\textbf{3.216E-02}  \\ ±9.549E-03\end{tabular} \\ \hline
D14           & \begin{tabular}[c]{@{}l@{}}6.493E-02  $\circ$\\ ±2.341E-02\end{tabular} & \begin{tabular}[c]{@{}l@{}}1.323E-01  $\downarrow$\\ ±5.574E-03\end{tabular} & \begin{tabular}[c]{@{}l@{}}\textbf{5.881E-02}  $^\circ$\\ ±2.029E-02\end{tabular} & \begin{tabular}[c]{@{}l@{}}7.499E-02  $\downarrow$\\ ±2.513E-02\end{tabular} & \begin{tabular}[c]{@{}l@{}}7.651E-02  $\downarrow$\\ ±2.350E-02\end{tabular} & \begin{tabular}[c]{@{}l@{}}1.309E-01 $\downarrow$\\ ±1.351E-02\end{tabular}  & \begin{tabular}[c]{@{}l@{}}6.048E-02  \\ ±2.226E-02\end{tabular} \\ \hline
D15        & \begin{tabular}[c]{@{}l@{}}1.376E-01  $\circ$\\ ±4.449E-02\end{tabular} & \begin{tabular}[c]{@{}l@{}}2.404E-01  $\downarrow$\\ ±8.469E-17\end{tabular} & \begin{tabular}[c]{@{}l@{}}1.509E-01  $\downarrow$\\ ±6.880E-02\end{tabular} & \begin{tabular}[c]{@{}l@{}}1.693E-01  $\downarrow$\\ ±7.097E-02\end{tabular} & \begin{tabular}[c]{@{}l@{}}1.870E-01  $\downarrow$\\ ±5.712E-02\end{tabular} & \begin{tabular}[c]{@{}l@{}}2.404E-01  $\downarrow$\\ ±8.469E-17\end{tabular} & \begin{tabular}[c]{@{}l@{}}\textbf{1.180E-01}  \\ ±4.739E-02\end{tabular} \\ \hline
R.       & 2.79 (10/4/1)                                                    & 4.19 (12/3/0)                                                    & 3.21 (9/6/0)                                                     & 3.72 (13/2/0)                                                    & 3.92 (13/2/0)                                                    & 4.81 (13/2/0)                                                    & \textbf{1.98  }                                                           \\ \hline
\end{tabular}
\end{table*}

\subsubsection {Analysis on Feature Selection Performance}
In addition to multi-objective evaluation metric, the minimum error (e.i., the first objective) for LMSSS and other large-scale feature selection algorithms on the test Pareto front has been presented in Table~\ref{tab:MCE}. As the final row shows, LMSSS achieved the top rank against all algorithms on dataset benchmarks. There is no significant loss in MCE for this algorithm. Multi-objective feature selection is a non-symmetric optimization problem, where classification error is usually more important and more difficult to minimize. Therefore, while the algorithm tries to decrease the number of features, achieving the highest classification accuracy is a serious challenge for this category of algorithms. This challenge is especially pronounced in large-scale feature selection, where finding a small set of features with high accuracy is crucial. From this table, LMSSS was more successful in this regard, finding more accurate feature subsets for the classification task. In some cases, such as the Lung and 11Tumor datasets, the difference between LMSSS and other algorithms is significantly large. This indicates the capability of the algorithm to handle an increasing number of features. 
Among the competitors, DAEA is in the second position with 3 ties and 12 losses compared to LMSSS. Additionally, as presented, the average standard deviation of the results from different runs using LMSSS is significantly less than that of other methods. In summery, LMSSS wins 78 and draws 12 out of the 90 comparisons for the test MCE results. The obvious reason behind this is the selection of top-ranked features in the first phase based on their correlation with the label. This results in removing a large set of irrelevant features and shrinking the search space for the optimizer to find a more accurate set of features in the second phase. The method's ability to consistently reduce the feature set while maintaining or improving classification accuracy underscores its efficiency in handling high-dimensional data. This can be evident from the superior performance of the algorithm on datasets like Lung and 11Tumor, especially as the number of features increases. Without that shrinkage, the optimizer would be incapable of finding the most relevant set to increase classification accuracy. 

\begin{table*}[]
\setlength{\tabcolsep}{2.4pt}
\centering
\caption{MCE values (the lower, the better) achieved by the competitor algorithms. The highest value in each dataset is highlighted in bold. The symbols $\uparrow$, $\downarrow$, and $^\circ$ indicate that LMSSS is significantly better, significantly worse, or not significantly different from the competitor algorithm, respectively. The last row displays the rank of each algorithm, with lower ranks indicating better performance, as determined by the Friedman test. Additionally, the number of wins/ties/losses of LMSSS compared to the other algorithms is shown in the final row.}
\label{tab:MCE}
\scriptsize
\begin{tabular}{llllllll}
\hline
       & DAEA                                                               & SparseEA                                                           & PMMOEA                                                             & SLMEA                                                              & SMMOEA                                                            & MSKEA                                                              & LMSSS                                                               \\ \hline
D1         & \begin{tabular}[c]{@{}l@{}}2.018E-01  $\downarrow$\\ ±7.454E-02\end{tabular}   & \begin{tabular}[c]{@{}l@{}}3.105E-01    $\downarrow$\\ ±6.679E-02\end{tabular} & \begin{tabular}[c]{@{}l@{}}2.421E-01   $\downarrow$\\  ±8.915E-02\end{tabular} & \begin{tabular}[c]{@{}l@{}}2.632E-01    $\downarrow$\\ ±8.407E-02\end{tabular} & \begin{tabular}[c]{@{}l@{}}2.579E-01   $\downarrow$\\ ±7.615E-02\end{tabular} & \begin{tabular}[c]{@{}l@{}}3.281E-01    $\downarrow$\\ ±4.301E-02\end{tabular} & \begin{tabular}[c]{@{}l@{}}\textbf{1.684E-01}   \\  ±3.497E-02\end{tabular} \\ \hline
D2         & \begin{tabular}[c]{@{}l@{}}8.395E-02   $\downarrow$\\ ±3.065E-02\end{tabular}  & \begin{tabular}[c]{@{}l@{}}1.136E-01  $\downarrow$\\ ±4.953E-02\end{tabular}   & \begin{tabular}[c]{@{}l@{}}8.395E-02  $\downarrow$\\ ±4.232E-02\end{tabular}   & \begin{tabular}[c]{@{}l@{}}1.025E-01  $\downarrow$\\ ±3.852E-02\end{tabular}   & \begin{tabular}[c]{@{}l@{}}9.506E-02  $\downarrow$\\ ±5.024E-02\end{tabular}  & \begin{tabular}[c]{@{}l@{}}1.358E-01  $\downarrow$\\ ±3.931E-02\end{tabular}   & \begin{tabular}[c]{@{}l@{}}\textbf{6.173E-02}  \\ ±2.971E-02\end{tabular}   \\ \hline
D3      & \begin{tabular}[c]{@{}l@{}}2.458E-01    $\downarrow$\\ ±5.365E-02\end{tabular} & \begin{tabular}[c]{@{}l@{}}2.500E-01  $\downarrow$\\ ±4.020E-02\end{tabular}   & \begin{tabular}[c]{@{}l@{}}2.344E-01  $\downarrow$\\ ±5.671E-02\end{tabular}   & \begin{tabular}[c]{@{}l@{}}2.542E-01  $\downarrow$\\ ±6.127E-02\end{tabular}   & \begin{tabular}[c]{@{}l@{}}2.302E-01  $\downarrow$\\ ±4.456E-02\end{tabular}  & \begin{tabular}[c]{@{}l@{}}2.313E-01 $^\circ$\\ ±5.293E-02\end{tabular}   & \begin{tabular}[c]{@{}l@{}}\textbf{2.073E-01}  \\ ±3.804E-02\end{tabular}   \\ \hline
D4        & \begin{tabular}[c]{@{}l@{}}3.074E-01   $\downarrow$\\ ±8.474E-02\end{tabular}  & \begin{tabular}[c]{@{}l@{}}3.167E-01  $\downarrow$\\ ±9.245E-02\end{tabular}   & \begin{tabular}[c]{@{}l@{}}2.741E-01 $^\circ$\\ ±8.114E-02\end{tabular}   & \begin{tabular}[c]{@{}l@{}}3.148E-01  $\downarrow$\\ ±9.380E-02\end{tabular}   & \begin{tabular}[c]{@{}l@{}}3.333E-01  $\downarrow$\\ ±8.754E-02\end{tabular}  & \begin{tabular}[c]{@{}l@{}}3.741E-01  $\downarrow$\\ ±7.137E-02\end{tabular}   & \begin{tabular}[c]{@{}l@{}}\textbf{2.444E-01}  \\ ±4.279E-02\end{tabular}   \\ \hline
D5  & \begin{tabular}[c]{@{}l@{}}7.246E-02  $^\circ$\\ ±4.615E-02\end{tabular} & \begin{tabular}[c]{@{}l@{}}9.855E-02  $\downarrow$\\ ±4.412E-02\end{tabular}   & \begin{tabular}[c]{@{}l@{}}6.232E-02 $^\circ$\\ ±4.935E-02\end{tabular}   & \begin{tabular}[c]{@{}l@{}}8.696E-02  $\downarrow$\\ ±5.822E-02\end{tabular}   & \begin{tabular}[c]{@{}l@{}}1.087E-01  $\downarrow$\\ ±4.673E-02\end{tabular}  & \begin{tabular}[c]{@{}l@{}}8.696E-02  $\downarrow$\\ ±3.787E-02\end{tabular}   & \begin{tabular}[c]{@{}l@{}}\textbf{5.507E-02}  \\ ±4.262E-02\end{tabular}   \\ \hline
D6           & \begin{tabular}[c]{@{}l@{}}2.842E-01   $^\circ$\\ ±9.734E-02\end{tabular} & \begin{tabular}[c]{@{}l@{}}3.351E-01  $\downarrow$\\ ±8.349E-02\end{tabular}   & \begin{tabular}[c]{@{}l@{}}3.123E-01  $\downarrow$\\ ±7.811E-02\end{tabular}   & \begin{tabular}[c]{@{}l@{}}3.158E-01 $\downarrow$\\ ±7.819E-02\end{tabular}    & \begin{tabular}[c]{@{}l@{}}3.211E-01  $\downarrow$\\ ±7.982E-02\end{tabular}  & \begin{tabular}[c]{@{}l@{}}4.158E-01  $\downarrow$\\ ±2.119E-02\end{tabular}   & \begin{tabular}[c]{@{}l@{}}\textbf{2.649E-01}  \\ ±6.256E-02\end{tabular}   \\ \hline
D7         & \begin{tabular}[c]{@{}l@{}}8.889E-02    $^\circ$\\ ±5.331E-02\end{tabular} & \begin{tabular}[c]{@{}l@{}}8.611E-02  $^\circ$\\ ±4.634E-02\end{tabular}   & \begin{tabular}[c]{@{}l@{}}8.472E-02  $^\circ$\\ ±6.140E-02\end{tabular}   & \begin{tabular}[c]{@{}l@{}}9.583E-02  $^\circ$\\ ±6.394E-02\end{tabular}   & \begin{tabular}[c]{@{}l@{}}1.069E-01  $^\circ$\\ ±6.536E-02\end{tabular}  & \begin{tabular}[c]{@{}l@{}}1.389E-01  $\downarrow$\\ ±4.422E-02\end{tabular}   & \begin{tabular}[c]{@{}l@{}}\textbf{7.639E-02}  \\ ±3.643E-02\end{tabular}   \\ \hline
D8 & \begin{tabular}[c]{@{}l@{}}6.812E-02    $\downarrow$\\ ±5.438E-02\end{tabular} & \begin{tabular}[c]{@{}l@{}}1.101E-01  $\downarrow$\\ ±3.562E-02\end{tabular}   & \begin{tabular}[c]{@{}l@{}}8.406E-02  $\downarrow$\\ ±5.586E-02\end{tabular}   & \begin{tabular}[c]{@{}l@{}}8.551E-02  $\downarrow$\\ ±4.193E-02\end{tabular}   & \begin{tabular}[c]{@{}l@{}}9.855E-02  $\downarrow$\\ ±5.224E-02\end{tabular}  & \begin{tabular}[c]{@{}l@{}}1.058E-01  $\downarrow$\\ ±2.952E-02\end{tabular}   & \begin{tabular}[c]{@{}l@{}}\textbf{3.478E-02}  \\ ±3.682E-02\end{tabular}   \\ \hline
D9      & \begin{tabular}[c]{@{}l@{}}1.638E-01    $\downarrow$\\ ±3.699E-02\end{tabular} & \begin{tabular}[c]{@{}l@{}}2.080E-01  $\downarrow$\\ ±3.867E-02\end{tabular}   & \begin{tabular}[c]{@{}l@{}}2.109E-01  $\downarrow$\\ ±4.452E-02\end{tabular}   & \begin{tabular}[c]{@{}l@{}}2.057E-01  $\downarrow$\\ ±3.677E-02\end{tabular}   & \begin{tabular}[c]{@{}l@{}}2.172E-01  $\downarrow$\\ ±3.436E-02\end{tabular}  & \begin{tabular}[c]{@{}l@{}}2.471E-01  $\downarrow$\\ ±3.824E-02\end{tabular}   & \begin{tabular}[c]{@{}l@{}}\textbf{1.287E-01}  \\ ±2.062E-02\end{tabular}   \\ \hline
D10      & \begin{tabular}[c]{@{}l@{}}1.720E-01    $\downarrow$\\ ±3.913E-02\end{tabular} & \begin{tabular}[c]{@{}l@{}}1.860E-01  $\downarrow$\\ ±3.662E-02\end{tabular}   & \begin{tabular}[c]{@{}l@{}}1.548E-01  $\downarrow$\\ ±3.919E-02\end{tabular}   & \begin{tabular}[c]{@{}l@{}}2.000E-01  $\downarrow$\\ ±3.919E-02\end{tabular}   & \begin{tabular}[c]{@{}l@{}}1.849E-01  $\downarrow$\\ ±4.314E-02\end{tabular}  & \begin{tabular}[c]{@{}l@{}}1.989E-01  $\downarrow$\\ ±3.501E-02\end{tabular}   & \begin{tabular}[c]{@{}l@{}}\textbf{1.312E-01}  \\ ±3.274E-02\end{tabular}   \\ \hline
D11      & \begin{tabular}[c]{@{}l@{}}2.781E-01    $\downarrow$\\ ±6.139E-02\end{tabular} & \begin{tabular}[c]{@{}l@{}}3.295E-01  $\downarrow$\\ ±3.879E-02\end{tabular}   & \begin{tabular}[c]{@{}l@{}}3.200E-01  $\downarrow$\\ ±6.205E-02\end{tabular}   & \begin{tabular}[c]{@{}l@{}}3.333E-01  $\downarrow$\\ ±6.157E-02\end{tabular}   & \begin{tabular}[c]{@{}l@{}}3.486E-01  $\downarrow$\\ ±7.919E-02\end{tabular}  & \begin{tabular}[c]{@{}l@{}}3.486E-01  $\downarrow$\\ ±3.630E-02\end{tabular}   & \begin{tabular}[c]{@{}l@{}}\textbf{2.448E-01}  \\ ±5.181E-02\end{tabular}   \\ \hline
D12       & \begin{tabular}[c]{@{}l@{}}2.098E-01    $\downarrow$\\ ±4.274E-02\end{tabular} & \begin{tabular}[c]{@{}l@{}}2.517E-01  $\downarrow$\\ ±5.331E-02\end{tabular}   & \begin{tabular}[c]{@{}l@{}}2.494E-01  $\downarrow$\\ ±5.137E-02\end{tabular}   & \begin{tabular}[c]{@{}l@{}}2.546E-01  $\downarrow$\\ ±4.406E-02\end{tabular}   & \begin{tabular}[c]{@{}l@{}}2.598E-01  $\downarrow$\\ ±3.997E-02\end{tabular}  & \begin{tabular}[c]{@{}l@{}}2.851E-01  $\downarrow$\\ ±5.596E-02\end{tabular}   & \begin{tabular}[c]{@{}l@{}}\textbf{1.483E-01}  \\ ±2.625E-02\end{tabular}   \\ \hline
D13          & \begin{tabular}[c]{@{}l@{}}9.471E-02    $\downarrow$\\ ±2.412E-02\end{tabular} & \begin{tabular}[c]{@{}l@{}}1.212E-01  $\downarrow$\\ ±2.552E-02\end{tabular}   & \begin{tabular}[c]{@{}l@{}}1.032E-01  $\downarrow$\\ ±2.457E-02\end{tabular}   & \begin{tabular}[c]{@{}l@{}}1.148E-01  $\downarrow$\\ ±3.220E-02\end{tabular}   & \begin{tabular}[c]{@{}l@{}}1.122E-01  $\downarrow$\\ ±3.173E-02\end{tabular}  & \begin{tabular}[c]{@{}l@{}}1.370E-01  $\downarrow$\\ ±3.295E-02\end{tabular}   & \begin{tabular}[c]{@{}l@{}}\textbf{6.138E-02}  \\ ±2.757E-02\end{tabular}   \\ \hline
D14           & \begin{tabular}[c]{@{}l@{}}\textbf{1.154E-01}    $^\circ$\\ ±4.403E-02\end{tabular} & \begin{tabular}[c]{@{}l@{}}2.679E-01  $\downarrow$\\ ±7.022E-03\end{tabular}   & \begin{tabular}[c]{@{}l@{}}1.244E-01  $^\circ$\\ ±3.736E-02\end{tabular}   & \begin{tabular}[c]{@{}l@{}}1.487E-01 $\downarrow$\\ ±5.508E-02\end{tabular}    & \begin{tabular}[c]{@{}l@{}}1.397E-01  $^\circ$\\ ±5.198E-02\end{tabular}  & \begin{tabular}[c]{@{}l@{}}2.641E-01  $\downarrow$\\ ±2.809E-02\end{tabular}   & \begin{tabular}[c]{@{}l@{}}1.256E-01  \\ ±3.903E-02\end{tabular}   \\ \hline
D15        & \begin{tabular}[c]{@{}l@{}}2.956E-01    $\downarrow$\\ ±4.853E-02\end{tabular} & \begin{tabular}[c]{@{}l@{}}4.000E-01  $\downarrow$\\ ±1.694E-16\end{tabular}   & \begin{tabular}[c]{@{}l@{}}3.100E-01  $\downarrow$\\ ±6.956E-02\end{tabular}   & \begin{tabular}[c]{@{}l@{}}3.289E-01  $\downarrow$\\ ±7.097E-02\end{tabular}   & \begin{tabular}[c]{@{}l@{}}3.467E-01 $\downarrow$\\ ±5.713E-02\end{tabular}   & \begin{tabular}[c]{@{}l@{}}4.000E-01  $\downarrow$\\ ±1.694E-16\end{tabular}   & \begin{tabular}[c]{@{}l@{}}\textbf{2.756E-01}  \\ ±5.099E-02\end{tabular}   \\ \hline
R.       & 2.79 (12/3/0)                                                      & 4.28 (14/1/0)                                                      & 3.13 (11/4/0)                                                      & 3.73 (14/1/0)                                                      & 3.99 (13/2/0)                                                     & 4.93 (14/1/0)                                                      & \textbf{1.76  }                                                             \\ \hline
\end{tabular}
\end{table*}


\subsubsection{Distributions of Non-dominated Solutions }
\begin{figure*}
\centering
\begin{tabular}{cccc}

      \includegraphics[width=0.23\linewidth]{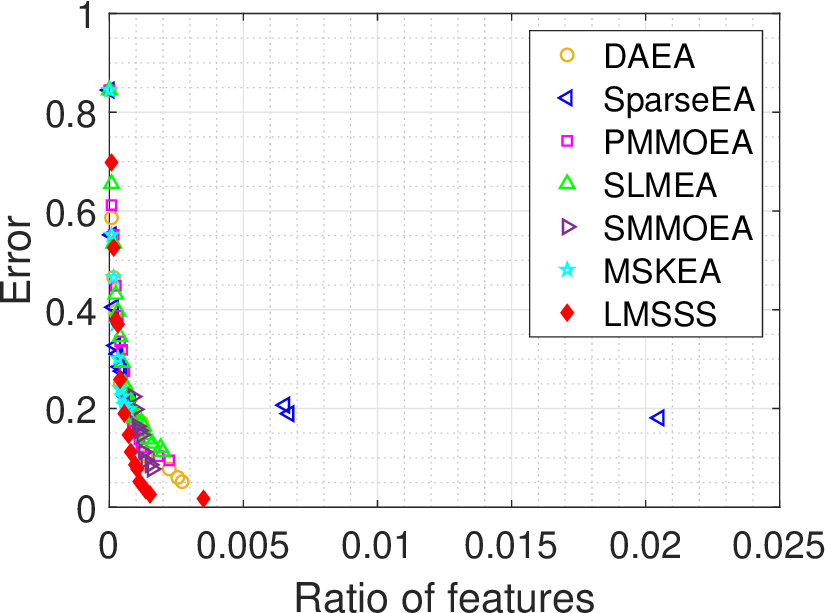} &
     
       \includegraphics[width=0.22\linewidth]{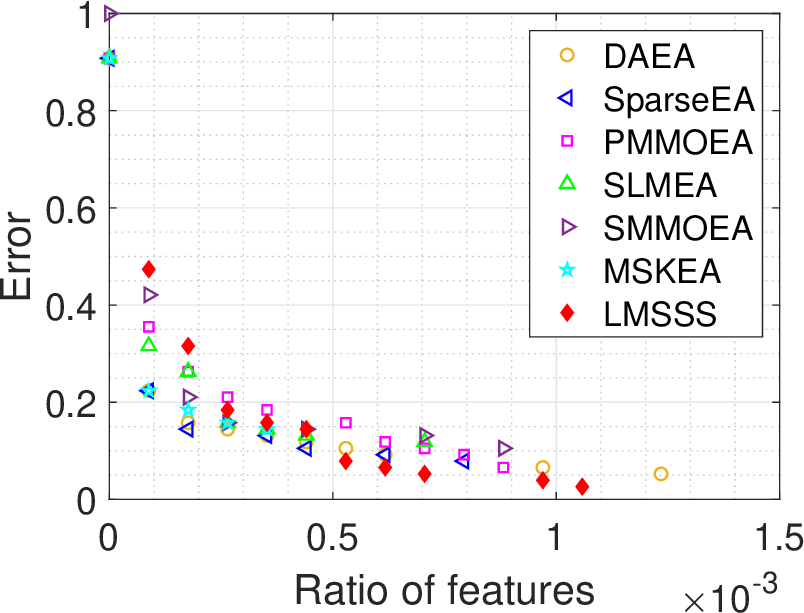} &
    
        \includegraphics[width=0.22\linewidth]{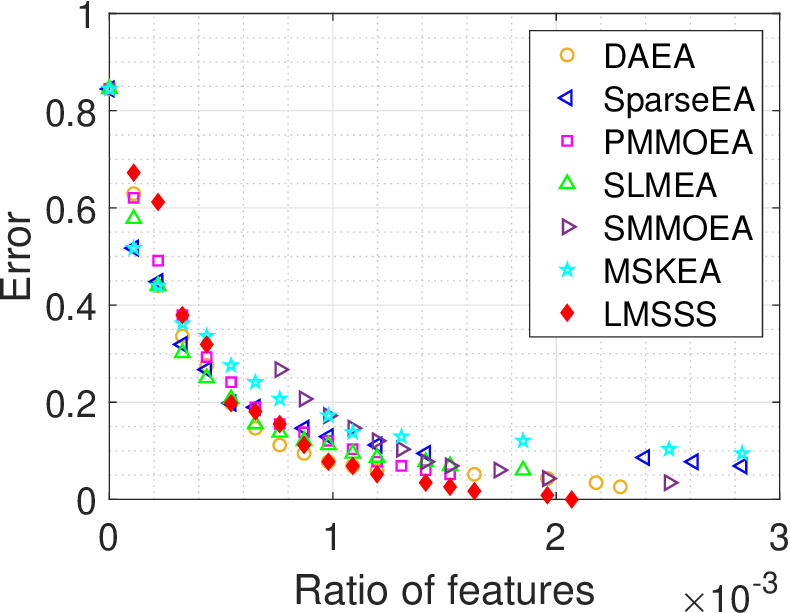} &
                \includegraphics[width=0.21\linewidth]{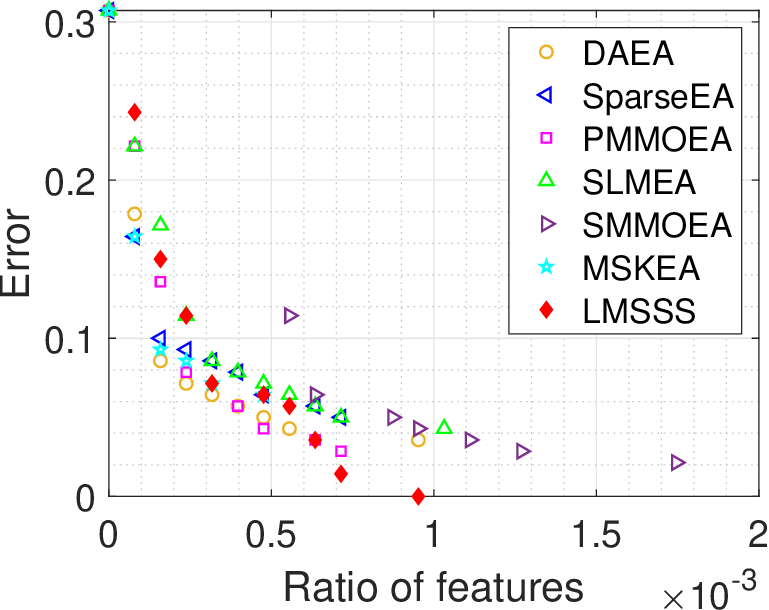}
                       \\
                       
                         \includegraphics[width=0.22\linewidth]{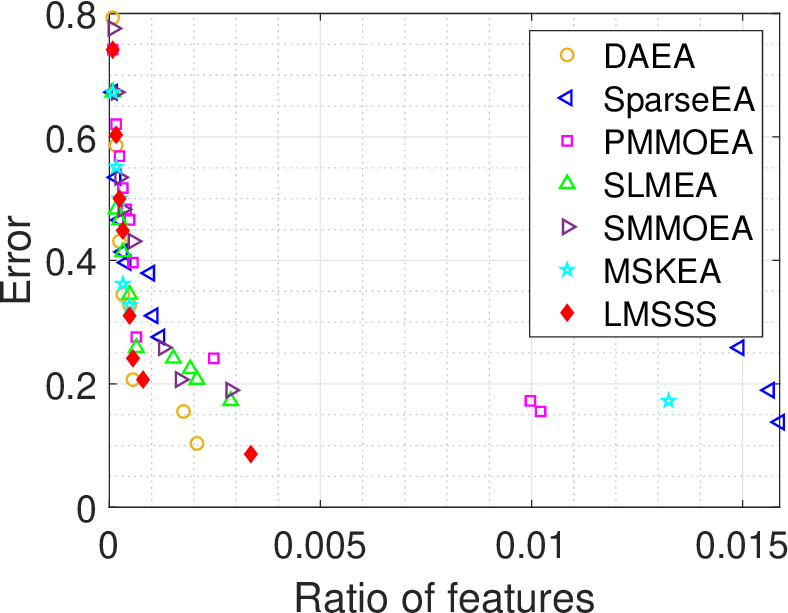} &
     
       \includegraphics[width=0.22\linewidth]{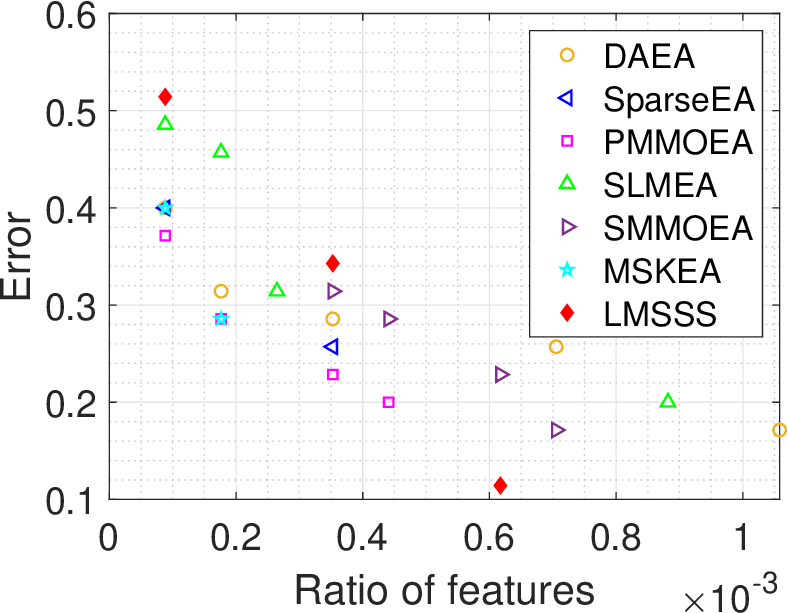} &
    
        \includegraphics[width=0.22\linewidth]{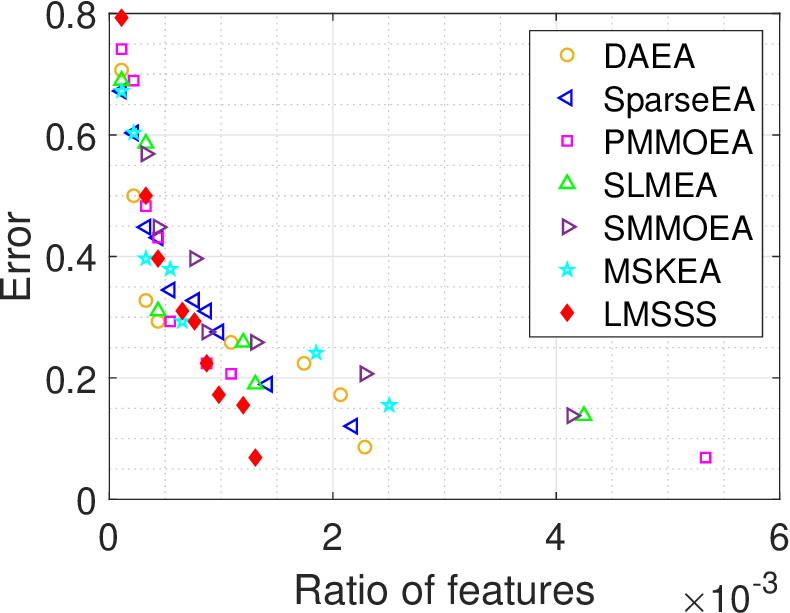} &
                \includegraphics[width=0.22\linewidth]{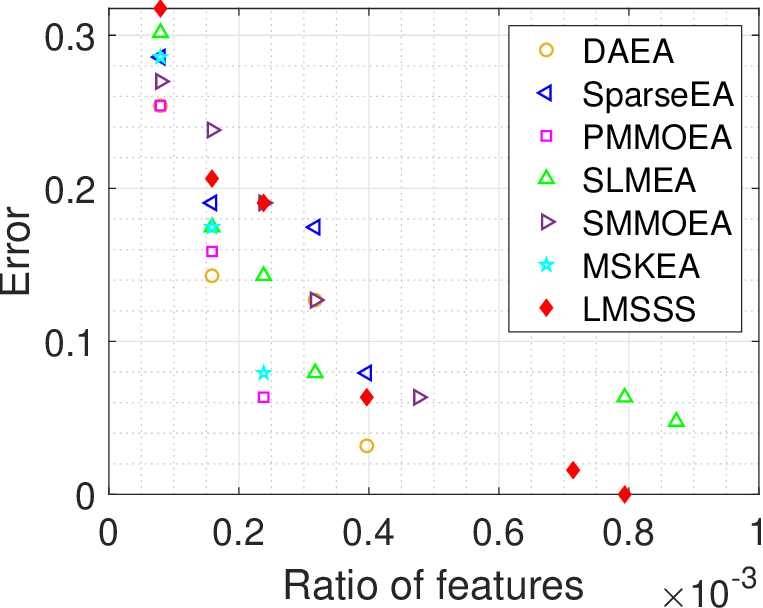}
                       \\11Tumor &CLL\_SUB&Carcinom&Lung\\

\end{tabular}
\caption{Top: resultant Pareto fronts from all competitive algorithms on some sample train sets, bottom: resultant Pareto fronts from all competitive algorithms on some sample test sets. }
\label{fig: PFs}
\end{figure*}

To provide an intuitive analysis, the distributions of the  non-dominated solutions from the  run with median HV of each algorithm are illustrated in Fig.~\ref{fig: PFs}. The first and second rows of these figures display all the non-dominated solutions for some sample datsets on the training sets and test sets, respectively. In other words, the solutions shown in the second row are derived from those in the first row with applying the NDS algorithm on resultant points. On the 11Tumor dataset, LMSSS significantly outperformed other algorithms by achieving a Pareto front on the training set with a very small set of features (i.e., less than 0.005 percent of the total number of features) and a very low classification error. Similarly, on the test set, the solution with the lowest error using a feature set comprising only 0.0035 percent of the total number of features belongs to LMSSS, while other datasets could not achieve this even with a higher number of features. Moreover, algorithms like SparseEA or PMMOEA reached the error of 0.2 with subsets sized between 0.01 to 0.015 percent of the total features, whereas LMSSS achieved the same error rate with only 0.001 percent of the features.

Similarly, on the CLL\_SUB dataset, the feature set with the minimum error belongs to LMSSS, using a set of only 6 features. In terms of the number of features, all algorithms achieved a similar range (i.e., 1 to 12 features out of 11,000 features). However, in terms of the accuracy of the selected features, LMSSS surpassed all algorithms, especially on the test set, with a significant margin. A similar pattern can be observed on the Carcinoma dataset, where the range of resultant feature subsets for most algorithms is from 1 to 26 features out of 9,182 total features, while LMSSS achieved the best classification accuracy with 19 features. None of the other algorithms could reduce the error as much. This is also evident on the test set, where LMSSS obtained a very small set of features in the range of 1 to 19 features with a minimum error of 5\%, whereas the other algorithms achieved an error of at least 20\% with this number of features.
The final illustrated Pareto front is on the Lung dataset, which has a total of 12,600 features. Although the distribution over the number of features is similar for all algorithms, LMSSS again obtained solutions with the minimum classification error using a set of 11 features. From the Pareto front, it is clear that for smaller sets (i.e., fewer than 8 features), other algorithms achieved better classification error.


\subsubsection{ Running Time Analysis}
\begin{figure*}
\centering
\begin{tabular}{cccc}

      \includegraphics[width=0.3\linewidth]{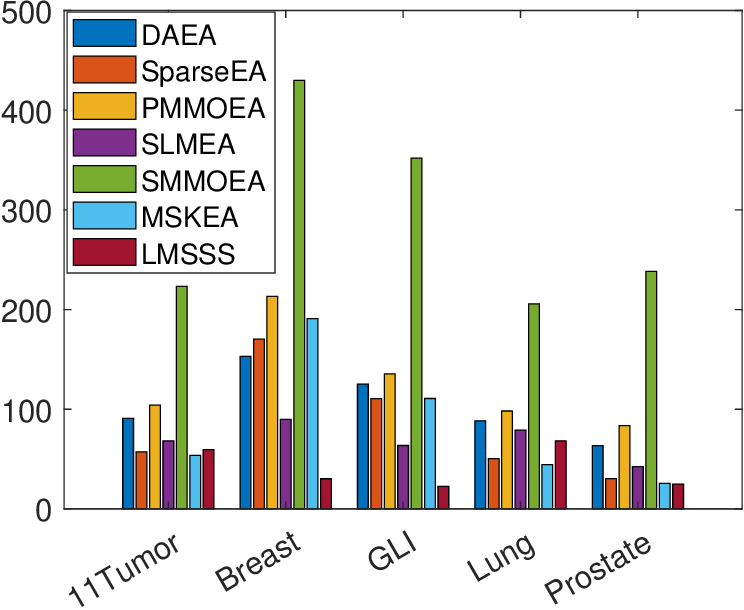} &
     
       \includegraphics[width=0.3\linewidth]{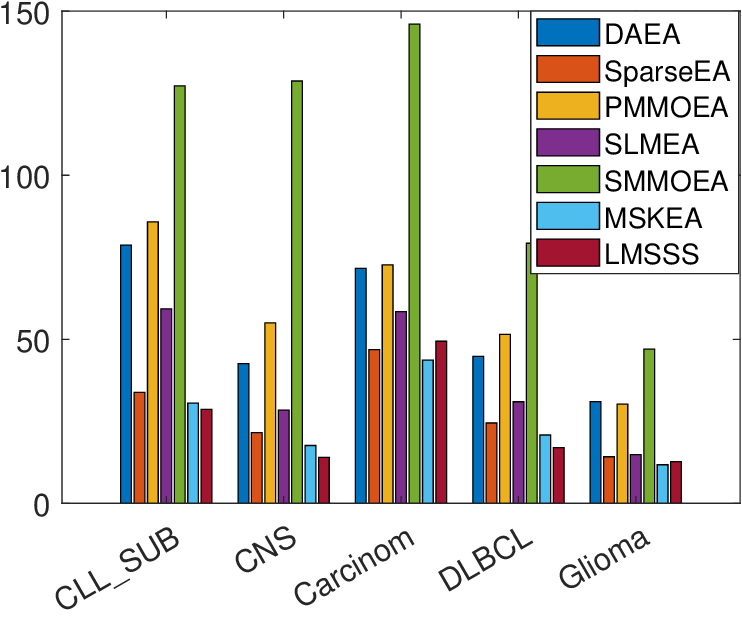} &
    
        \includegraphics[width=0.3\linewidth]{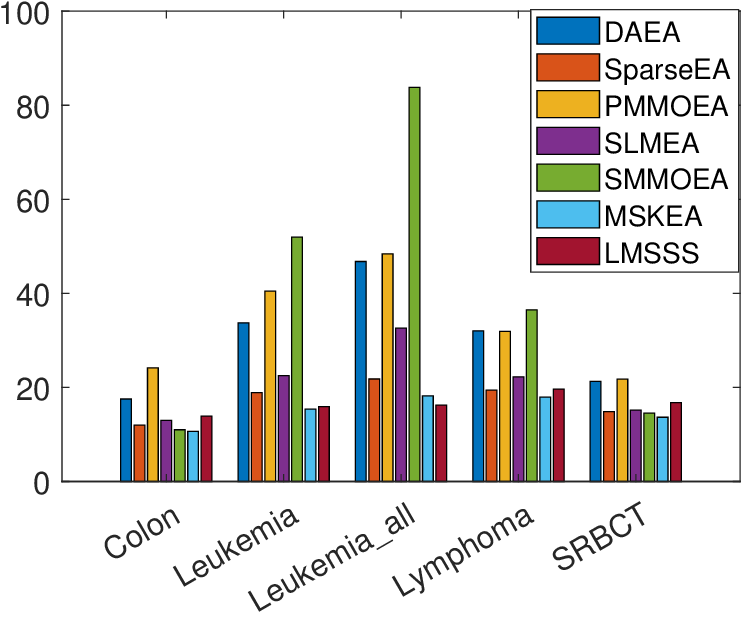} 
               
                       \\

\end{tabular}
\caption{The running time of a single run for each algorithm across different datasets. The times are measured in seconds. Each bar represents the running time of one algorithm for each dataset on the x-axis.  }
\label{fig:time}
\end{figure*}

Fig.~\ref{fig:time} shows the time consumed by each algorithm on different datasets. As these algorithms are designed for large-scale feature selection, they are tailored to solve large-scale problems, which can be very time-consuming. Therefore, analyzing the time required is crucial. As illustrated, in most cases, LMSSS consumes less time than all other algorithms. The obvious reason, as discussed earlier, is the two-phase scheme in which many irrelevant features are eliminated in the first phase. The time-consuming aspect of these algorithms is the fitness evaluation (i.e., classification). In LMSSS, the shrunk space allows for evaluating a smaller set of features, resulting in less overall time. It is worth noting that the first phase is less computationally expensive and does not significantly affect the overall time. However, all phases have been considered in the time calculation. Thus, shrinking the search space in large-scale feature selection is not only efficient in finding more accurate solutions but also accelerates the convergence of the algorithm. In this ranking, MSKEA is in the second position, while SMMOEA is in the last position, resulting in very long run times.

\begin{table*}[]
\centering 
\caption{HV  values for component contribution analysis. Each column represents the results of an algorithm obtained by adding a component to the algorithm in the previous column, demonstrating the contribution of the added component. The schemes include: InitNSGA-II (only initialization added to NSGA-II), SS-NSGA-II (InitNSGA-II + search space shrinking), and LMSSS (SS-NSGA-II + New operators). The symbols $\uparrow$, $\downarrow$, and $^\circ$ indicate that the schema on corresponding column is significantly better, significantly worse, or not significantly different from the previous scheme, respectively. Additionally, the number of wins/ties/losses of LMSSS compared to the other schemes is shown in the final row.  }
\label{tab:Contribution_HV}
\scriptsize
\begin{tabular}{lllll}
\hline
              & \multicolumn{4}{c}{HV}                                                                                                                                                                                                                                                   \\ \hline
       & NSGA-II                                                           & Init-NSGA-II                                                      & SS-NSGA-II                                                         & LMSSS                                                             \\ \hline
D1         & \begin{tabular}[c]{@{}l@{}}6.924E-01  \\ ±4.021E-02\end{tabular} & \begin{tabular}[c]{@{}l@{}}6.838E-01 $^\circ$ \\  ±7.427E-02\end{tabular} & \begin{tabular}[c]{@{}l@{}}7.644E-01 $\uparrow$ \\ ±4.521E-02\end{tabular} & \begin{tabular}[c]{@{}l@{}}\textbf{8.305E-01} $\uparrow$ \\ ±3.491E-02\end{tabular} \\ \hline
D2         & \begin{tabular}[c]{@{}l@{}}8.789E-01  \\ ±1.265E-02\end{tabular} & \begin{tabular}[c]{@{}l@{}}9.128E-01 $\uparrow$\\  ±2.801E-02\end{tabular} & \begin{tabular}[c]{@{}l@{}}\textbf{9.413E-01}  $\uparrow$\\ ±2.861E-02\end{tabular} & \begin{tabular}[c]{@{}l@{}}9.370E-01 $^\circ$ \\ ±2.956E-02\end{tabular} \\ \hline
D3      & \begin{tabular}[c]{@{}l@{}}7.085E-01  \\ ±1.716E-02\end{tabular} & \begin{tabular}[c]{@{}l@{}}7.756E-01 $\uparrow$ \\ ±3.764E-02\end{tabular} & \begin{tabular}[c]{@{}l@{}}7.632E-01 $^\circ$ \\ ±6.489E-02\end{tabular} & \begin{tabular}[c]{@{}l@{}}\textbf{7.921E-01} $^\circ$ \\ ±3.794E-02\end{tabular} \\ \hline
D4        & \begin{tabular}[c]{@{}l@{}}5.647E-01  \\ ±3.044E-02\end{tabular} & \begin{tabular}[c]{@{}l@{}}7.201E-01  $\uparrow$\\ ±5.551E-02\end{tabular} & \begin{tabular}[c]{@{}l@{}}6.794E-01 $^\circ$ \\ ±1.277E-01\end{tabular} & \begin{tabular}[c]{@{}l@{}}\textbf{7.551E-01} $^\circ$ \\ ±4.272E-02\end{tabular} \\ \hline
D5      & \begin{tabular}[c]{@{}l@{}}6.531E-01  \\ ±3.448E-02\end{tabular} & \begin{tabular}[c]{@{}l@{}}9.027E-01  $\uparrow$\\ ±6.323E-02\end{tabular} & \begin{tabular}[c]{@{}l@{}}8.824E-01 $^\circ$ \\ ±7.050E-02\end{tabular} & \begin{tabular}[c]{@{}l@{}}\textbf{9.446E-01} $\uparrow$ \\ ±4.258E-02\end{tabular} \\ \hline
D6           & \begin{tabular}[c]{@{}l@{}}5.289E-01  \\ ±1.592E-02\end{tabular} & \begin{tabular}[c]{@{}l@{}}\textbf{7.419E-01} $\uparrow$ \\ ±7.091E-02\end{tabular} & \begin{tabular}[c]{@{}l@{}}7.157E-01 $^\circ$ \\ ±6.713E-02\end{tabular} & \begin{tabular}[c]{@{}l@{}}7.348E-01 $^\circ$ \\ ±6.249E-02\end{tabular} \\ \hline
D7         & \begin{tabular}[c]{@{}l@{}}6.817E-01  \\ ±3.556E-02\end{tabular} & \begin{tabular}[c]{@{}l@{}}8.914E-01 $\uparrow$ \\ ±6.791E-02\end{tabular} & \begin{tabular}[c]{@{}l@{}}9.373E-01 $^\circ$ \\ ±6.164E-02\end{tabular} & \begin{tabular}[c]{@{}l@{}}\textbf{9.233E-01} $^\circ$ \\ ±3.645E-02\end{tabular} \\ \hline
D8 & \begin{tabular}[c]{@{}l@{}}7.071E-01  \\ ±1.385E-02\end{tabular} & \begin{tabular}[c]{@{}l@{}}8.912E-01 $\uparrow$ \\ ±4.809E-02\end{tabular} & \begin{tabular}[c]{@{}l@{}}9.404E-01 $\uparrow$ \\ ±7.081E-02\end{tabular} & \begin{tabular}[c]{@{}l@{}}\textbf{9.649E-01} $^\circ$ \\ ±3.680E-02\end{tabular} \\ \hline
D9      & \begin{tabular}[c]{@{}l@{}}5.705E-01  \\ ±1.168E-02\end{tabular} & \begin{tabular}[c]{@{}l@{}}8.310E-01 $\uparrow$ \\ ±3.052E-02\end{tabular} & \begin{tabular}[c]{@{}l@{}}8.365E-01 $^\circ$ \\ ±3.092E-02\end{tabular} & \begin{tabular}[c]{@{}l@{}}\textbf{8.708E-01} $\uparrow$ \\ ±2.057E-02\end{tabular} \\ \hline
D10      & \begin{tabular}[c]{@{}l@{}}5.549E-01  \\ ±1.092E-02\end{tabular} & \begin{tabular}[c]{@{}l@{}}8.375E-01 $\uparrow$ \\ ±3.336E-02\end{tabular} & \begin{tabular}[c]{@{}l@{}}8.160E-01 $^\circ$\\  ±6.232E-02\end{tabular} & \begin{tabular}[c]{@{}l@{}}\textbf{8.686E-01} $\uparrow$ \\ ±3.272E-02\end{tabular} \\ \hline
D11     & \begin{tabular}[c]{@{}l@{}}4.390E-01  \\ ±2.142E-02\end{tabular} & \begin{tabular}[c]{@{}l@{}}7.150E-01 $\uparrow$ \\ ±5.060E-02\end{tabular} & \begin{tabular}[c]{@{}l@{}}7.294E-01  \\ ±3.950E-02\end{tabular} & \begin{tabular}[c]{@{}l@{}}\textbf{7.551E-01} $^\circ$ \\ ±5.179E-02\end{tabular} \\ \hline
D12       & \begin{tabular}[c]{@{}l@{}}5.208E-01 \\  ±1.063E-02\end{tabular} & \begin{tabular}[c]{@{}l@{}}8.150E-01 $\uparrow$ \\ ±4.529E-02\end{tabular} & \begin{tabular}[c]{@{}l@{}}8.170E-01 $^\circ$ \\ ±3.607E-02\end{tabular} & \begin{tabular}[c]{@{}l@{}}\textbf{8.513E-01} $\uparrow$ \\ ±2.616E-02\end{tabular} \\ \hline
D13          & \begin{tabular}[c]{@{}l@{}}5.868E-01  \\ ±5.691E-03\end{tabular} & \begin{tabular}[c]{@{}l@{}}9.007E-01 $\uparrow$ \\ ±2.586E-02\end{tabular} & \begin{tabular}[c]{@{}l@{}}9.221E-01 $\uparrow$ \\ ±2.406E-02\end{tabular} & \begin{tabular}[c]{@{}l@{}}\textbf{9.384E-01} $^\circ$ \\ ±2.758E-02\end{tabular} \\ \hline
D14           & \begin{tabular}[c]{@{}l@{}}5.749E-01  \\ ±1.268E-02\end{tabular} & \begin{tabular}[c]{@{}l@{}}\textbf{8.846E-01} $\uparrow$ \\ ±4.128E-02\end{tabular} & \begin{tabular}[c]{@{}l@{}}8.359E-01 $\downarrow$ \\ ±2.659E-02\end{tabular} & \begin{tabular}[c]{@{}l@{}}8.743E-01  $\uparrow$\\ ±3.903E-02\end{tabular} \\ \hline
D15        & \begin{tabular}[c]{@{}l@{}}4.129E-01  \\ ±1.958E-02\end{tabular} & \begin{tabular}[c]{@{}l@{}}7.130E-01 $\uparrow$ \\ ±4.900E-02\end{tabular} & \begin{tabular}[c]{@{}l@{}}7.200E-01 $^\circ$ \\ ±6.348E-02\end{tabular} & \begin{tabular}[c]{@{}l@{}}\textbf{7.244E-01} $^\circ$ \\ ±5.099E-02\end{tabular} \\ \hline
R.       & 4.36 (15/0/0)                                                            & 2.74 (13/0/2)                                                            & 2.57 (14/0/1)                                                             & \textbf{2.19}                                                             \\ \hline
\end{tabular}

\end{table*}

\begin{table*}[]
\centering 
\caption{IGD values for component contribution analysis. Each column represents the results of an algorithm obtained by adding a component to the algorithm in the previous column, demonstrating the contribution of the added component. The schemes include: InitNSGA-II (only initialization added to NSGA-II), SS-NSGA-II (InitNSGA-II + search space shrinking), and LMSSS (SS-NSGA-II + New operators). The symbols $\uparrow$, $\downarrow$, and $^\circ$ indicate that the schema on corresponding column is significantly better, significantly worse, or not significantly different from the previous scheme, respectively. Additionally, the number of wins/ties/losses of LMSSS compared to the other schemes is shown in the final row.  }
\label{tab:Contribution_IGD}
\scriptsize
\begin{tabular}{lllll}
\hline
              & \multicolumn{4}{c}{IGD}                                                                                                                                                                                                                                                     \\ \hline
       & NSGA-II                                                             & Init-NSGA-II                                                      & SS-NSGA-II                                                         & LMSSS                                                             \\ \hline
D1            & \begin{tabular}[c]{@{}l@{}}1.337E-01  \\ ±2.035E-02\end{tabular}   & \begin{tabular}[c]{@{}l@{}}1.625E-01 $^\circ$ \\ ±5.595E-02\end{tabular} & \begin{tabular}[c]{@{}l@{}}8.178E-02 $\uparrow$ \\ ±3.840E-02\end{tabular} & \begin{tabular}[c]{@{}l@{}}\textbf{4.422E-02} $\uparrow$ \\ ±1.681E-02\end{tabular} \\ \hline
D2            & \begin{tabular}[c]{@{}l@{}}1.810E-01    \\ ±1.603E-02\end{tabular} & \begin{tabular}[c]{@{}l@{}}5.899E-02 $\uparrow$ \\ ±1.813E-02\end{tabular} & \begin{tabular}[c]{@{}l@{}}\textbf{3.011E-02} $\uparrow$ \\ ±2.205E-02\end{tabular} & \begin{tabular}[c]{@{}l@{}}5.245E-02 $\downarrow$ \\ ±2.227E-02\end{tabular} \\ \hline
D3            & \begin{tabular}[c]{@{}l@{}}2.020E-01    \\ ±4.989E-03\end{tabular} & \begin{tabular}[c]{@{}l@{}}1.048E-01 $\uparrow$ \\ ±2.466E-02\end{tabular} & \begin{tabular}[c]{@{}l@{}}1.189E-01 $^\circ$ \\ ±4.833E-02\end{tabular} & \begin{tabular}[c]{@{}l@{}}\textbf{9.839E-02} $^\circ$ \\ ±1.959E-02\end{tabular} \\ \hline
D4            & \begin{tabular}[c]{@{}l@{}}2.708E-01   \\  ±2.107E-02\end{tabular} & \begin{tabular}[c]{@{}l@{}}1.474E-01 $\uparrow$ \\ ±5.168E-02\end{tabular} & \begin{tabular}[c]{@{}l@{}}1.889E-01 $^\circ$ \\ ±1.243E-01\end{tabular} & \begin{tabular}[c]{@{}l@{}}\textbf{1.178E-01} $^\circ$ \\ ±3.596E-02\end{tabular} \\ \hline
D5            & \begin{tabular}[c]{@{}l@{}}2.413E-01    \\ ±6.917E-03\end{tabular} & \begin{tabular}[c]{@{}l@{}}7.890E-02 $\uparrow$ \\ ±2.835E-02\end{tabular} & \begin{tabular}[c]{@{}l@{}}8.488E-02 $^\circ$ \\ ±2.443E-02\end{tabular} & \begin{tabular}[c]{@{}l@{}}\textbf{5.302E-02} $\uparrow$ \\ ±3.151E-02\end{tabular} \\ \hline
D6            & \begin{tabular}[c]{@{}l@{}}3.014E-01    \\ ±8.680E-03\end{tabular} & \begin{tabular}[c]{@{}l@{}}\textbf{1.074E-01} $\uparrow$ \\ ±5.588E-02\end{tabular} & \begin{tabular}[c]{@{}l@{}}1.294E-01 $^\circ$ \\ ±5.634E-02\end{tabular} & \begin{tabular}[c]{@{}l@{}}1.111E-01 $^\circ$ \\ ±5.037E-02\end{tabular} \\ \hline
D7            & \begin{tabular}[c]{@{}l@{}}2.785E-01    \\ ±5.681E-03\end{tabular} & \begin{tabular}[c]{@{}l@{}}8.082E-02 $\uparrow$ \\ ±3.844E-02\end{tabular} & \begin{tabular}[c]{@{}l@{}}6.815E-02 $\uparrow$ \\ ±3.383E-02\end{tabular} & \begin{tabular}[c]{@{}l@{}}\textbf{6.220E-02} $^\circ$ \\ ±2.298E-02\end{tabular} \\ \hline
D8            & \begin{tabular}[c]{@{}l@{}}2.856E-01    \\ ±1.037E-02\end{tabular} & \begin{tabular}[c]{@{}l@{}}8.314E-02 $\uparrow$ \\ ±2.053E-02\end{tabular} & \begin{tabular}[c]{@{}l@{}}8.286E-02 $^\circ$ \\ ±1.722E-02\end{tabular} & \begin{tabular}[c]{@{}l@{}}\textbf{5.372E-02} $\uparrow$ \\ ±2.107E-02\end{tabular} \\ \hline
D9            & \begin{tabular}[c]{@{}l@{}}3.041E-01    \\ ±3.069E-03\end{tabular} & \begin{tabular}[c]{@{}l@{}}3.231E-02 $\uparrow$ \\ ±1.644E-02\end{tabular} & \begin{tabular}[c]{@{}l@{}}3.069E-02 $^\circ$ \\ ±1.213E-02\end{tabular} & \begin{tabular}[c]{@{}l@{}}\textbf{2.091E-02} $\uparrow$ \\ ±7.557E-03\end{tabular} \\ \hline
D10           & \begin{tabular}[c]{@{}l@{}}3.227E-01    \\ ±4.925E-03\end{tabular} & \begin{tabular}[c]{@{}l@{}}7.461E-02  $\uparrow$\\ ±1.503E-02\end{tabular} & \begin{tabular}[c]{@{}l@{}}7.975E-02 $^\circ$ \\ ±2.638E-02\end{tabular} & \begin{tabular}[c]{@{}l@{}}\textbf{6.281E-02} $\uparrow$ \\ ±2.393E-02\end{tabular} \\ \hline
D11           & \begin{tabular}[c]{@{}l@{}}3.788E-01    \\ ±1.532E-02\end{tabular} & \begin{tabular}[c]{@{}l@{}}1.217E-01 $\uparrow$ \\ ±4.603E-02\end{tabular} & \begin{tabular}[c]{@{}l@{}}1.066E-01 $^\circ$ \\ ±3.665E-02\end{tabular} & \begin{tabular}[c]{@{}l@{}}\textbf{9.043E-02} $^\circ$ \\ ±3.311E-02\end{tabular} \\ \hline
D12           & \begin{tabular}[c]{@{}l@{}}3.534E-01    \\ ±3.480E-03\end{tabular} & \begin{tabular}[c]{@{}l@{}}5.374E-02 $\uparrow$ \\ ±2.036E-02\end{tabular} & \begin{tabular}[c]{@{}l@{}}3.778E-02  $\uparrow$\\ ±1.276E-02\end{tabular} & \begin{tabular}[c]{@{}l@{}}\textbf{2.553E-02} $\uparrow$ \\ ±6.114E-03\end{tabular} \\ \hline
D13           & \begin{tabular}[c]{@{}l@{}}3.265E-01    \\ ±3.088E-03\end{tabular} & \begin{tabular}[c]{@{}l@{}}5.596E-02  $\uparrow$\\ ±1.834E-02\end{tabular} & \begin{tabular}[c]{@{}l@{}}3.996E-02 $\uparrow$ \\ ±1.022E-02\end{tabular} & \begin{tabular}[c]{@{}l@{}}\textbf{3.216E-02} $\uparrow$ \\ ±9.549E-03\end{tabular} \\ \hline
D14           & \begin{tabular}[c]{@{}l@{}}3.803E-01    \\ ±3.955E-03\end{tabular} & \begin{tabular}[c]{@{}l@{}}6.582E-02 $\uparrow$ \\ ±2.223E-02\end{tabular} & \begin{tabular}[c]{@{}l@{}}8.021E-02 $^\circ$ \\ ±1.385E-02\end{tabular} & \begin{tabular}[c]{@{}l@{}}\textbf{6.048E-02} $^\circ$ \\ ±2.226E-02\end{tabular} \\ \hline
D15           & \begin{tabular}[c]{@{}l@{}}4.176E-01    \\ ±1.335E-02\end{tabular} & \begin{tabular}[c]{@{}l@{}}1.271E-01 $\uparrow$ \\ ±4.740E-02\end{tabular} & \begin{tabular}[c]{@{}l@{}}1.228E-01 $^\circ$ \\ ±6.000E-02\end{tabular} & \begin{tabular}[c]{@{}l@{}}\textbf{1.180E-01}  $^\circ$\\ ±4.739E-02\end{tabular} \\ \hline
R.       & 3.45 (15/0/0)                                                              & 1.81 (14/0/1)                                                             & 1.67 (14/0/1)                                                             & \textbf{1.21}                                                             \\ \hline
\end{tabular}

\end{table*}
\subsubsection{Analysis on Component Contribution}
LMSSS consists of three main components: the initialization strategy, the shrinking phase and the binary crossover and mutation operators. Assessing the effectiveness of each component is crucial. To demonstrate this, three variant algorithms  were created to highlight the contributions of these three components. Each time a new component will be added to the previous version.   With these variations, the impact of each component can be investigated by checking the improvement of each algorithm compared to the previous variant. Tables~\ref{tab:Contribution_HV}, \ref{tab:Contribution_IGD} and \ref{tab:Contribution_MCE} present the values of HV, IGD, and MCE for all these variants. The last rows represent the ranking across all variants. For each column, the comparison markers indicate the significance of the corresponding algorithm compared to the previous column, highlighting the efficiency of adding the corresponding component.

\textbf{NSGA-II:} As the selection scheme of the proposed method is similar to NSGA-II, the base algorithm is the original version of NSGA-II, utilizing uniform crossover and bit flip mutation. This algorithm is applied to the entire feature set, without any shrinking strategy. Initialization follows the standard uniform method, generating an equal number of 0s and 1s in each individual, resulting in the selection of almost 50\% of the total features per individual. This preserves the individual-level uniformity. 
  
   \textbf{Init-NSGA-II:} The second version incorporates an initialization strategy into NSGA-II to investigate the impact of the initialization component. For each individual, a random integer number from 1 to the total number of features is selected to represent the number of chosen features. This approach ensures uniformity at both the individual and population levels. 
  From Tables~\ref{tab:Contribution_HV} and \ref{tab:Contribution_IGD}  it is evident that initialization has a significant impact on the HV and IGD values compared to NSGA-II. However, many research studies underestimate the importance of this component in sparse large-scale optimization problems. Even when employing a more uniform initialization, its contribution is often overlooked as a critical component.
Uniformity at the population level results in smaller feature sets in the initial population, which helps the optimizer start with fewer features. In contrast, with regular initialization, the optimizer expends a significant amount of effort to reduce the number of features. As previously noted, in many sparse large-scale datasets, an accurate set of features often consists of a very small subset. Therefore, providing smaller feature sets in the initial population is a substantial advantage for the optimizer.
The impact of initialization is also evident on the MCE. From Table \ref{tab:Contribution_MCE}, Init-NSGA-II outperformed NSGA-II in 7 out of 15 datasets, with 4 ties and 4 losses. This is an interesting finding because one might initially assume that providing smaller feature sets in the initial population primarily benefits the number of features objective, thereby improving multi-objective evaluation measures such as HV and IGD. However, the results indicate that this initialization strategy also benefits the first objective, classification accuracy.
  
  \textbf{SS-NSGA-II:} This variant incorportaes the shrinking strategy to the Init-NSGA-II resulting in a smaller search space for the multi-objective feature selection algorithm. The top ranked features in terms of the correlation with the class label are selected for the second phase. 
  The shrinking phase improves HV and IGD in general, resulting in a higher rank for SS-NSGA-II compared to Init-NSGA-II as indicated in Table~\ref{tab:Contribution_HV_IGD}. The top selected features facilitate the search process for the optimizer, resulting in a smaller set of features with lower classification error. As Table~\ref{tab:Contribution_MCE} indicates, the classification error decreases in five datasets. For instance, the highly correlated features achieved an average classification error of 6\% on the DLBCL dataset, while the previous variant achieved an error of 10\% on this dataset. This improvement leads to better multi-objective evaluation measures, including HV and IGD. This strategy is beneficial for the optimizer as it avoids wasting time on irrelevant features in a vast search space. Increasing the number of features can exponentially increase the size of the search space. Hence, this strategy can significantly remove some regions for the optimizer to focus on more promising areas. In summary, the shrinking strategy contributes to the improvement of both objectives by removing some features to result in a smaller set of features in the new search space and eliminating irrelevant features to the class label, thereby increasing classification accuracy.
  
  \textbf{LMSSS:} The final variant, introduced as the proposed method, incorporates all components by adding the proposed generative operators to SS-NSGA-II. As discussed previously, one of the challenges of multi-objective feature selection is the inability of generative operators to build a diverse population. Consequently, in many optimization processes, after a few iterations, population diversity drops, leading to premature convergence. The crucial impact of generative operators is evident from the tables. Compared to the previous variant, changing the common genetic operators to the newly proposed operators results in higher IGD and HV values, representing diverse and high-quality feature subsets in the resultant Pareto front. Overall, this change increased the rank of the final proposed algorithm compared to the previous variant. In terms of MCE, as shown in the table, the new crossover and mutation operators result in lower error on 8 datasets, with no increase in any dataset. This highlights the key role of generative operators in large-scale optimization problems, significantly contributing to the classification error objective.

Overall, the final variant, LMSSS, integrating all components, achieved the lowest error on 13 out of 15 datasets compared to all variations. Therefore, the joint contribution of all components results in more accurate feature subsets in terms of classification error. Additionally, as the multi-metric evaluation metrics indicate, LMSSS outperformed all other variations with 13 wins out of 15 in terms of both HV and IGD metrics. This indicates that each component can help improve one or both objectives.

\begin{table}[]
\centering
\caption{MCE values for component contribution analysis. Each column represents the results of an algorithm obtained by adding a component to the algorithm in the previous column, demonstrating the contribution of the added component. The schemes include: InitNSGA-II (only initialization added to NSGA-II), MNSGA-II (InitNSGA-II + Multi-phase shrinking), and LMSSS (MNSGA-II + New operators). The symbols $\uparrow$, $\downarrow$, and $^\circ$ indicate that the schema on corresponding column is significantly better, significantly worse, or not significantly different from the previous scheme, respectively. Additionally, the number of wins/ties/losses of LMSSS compared to the other schemes is shown in the final row.}
\label{tab:Contribution_MCE}
\setlength{\tabcolsep}{4pt}
\scriptsize
\begin{tabular}{lllll}
\hline
\multicolumn{5}{c}{MCE}                                                                                                                                                                                                                                                                     \\ \hline
       & NSGA-II                                                             & Init-NSGA-II                                                      & SS-NSGA-II                                                         & LMSSS                                                             \\ \hline
D1         & \begin{tabular}[c]{@{}l@{}}2.386E-01  \\ ±4.388E-02\end{tabular}   & \begin{tabular}[c]{@{}l@{}}3.158E-01 $\downarrow$ \\ ±7.443E-02\end{tabular} & \begin{tabular}[c]{@{}l@{}}2.351E-01  $\uparrow$\\ ±4.528E-02\end{tabular} & \begin{tabular}[c]{@{}l@{}}\textbf{1.684E-01}  $\uparrow$ \\ ±3.497E-02\end{tabular} \\ \hline

D2         & \begin{tabular}[c]{@{}l@{}}4.938E-03    \\ ±1.303E-02\end{tabular} & \begin{tabular}[c]{@{}l@{}}8.642E-02  $\downarrow$\\ ±2.808E-02\end{tabular} & \begin{tabular}[c]{@{}l@{}}\textbf{5.802E-02} $\uparrow$ \\ ±2.866E-02\end{tabular} & \begin{tabular}[c]{@{}l@{}}6.173E-02 $^\circ$ \\ ±2.971E-02\end{tabular} \\ \hline

D3      & \begin{tabular}[c]{@{}l@{}}1.396E-01    \\ ±2.000E-02\end{tabular} & \begin{tabular}[c]{@{}l@{}}2.240E-01 $\downarrow$ \\ ±3.768E-02\end{tabular} & \begin{tabular}[c]{@{}l@{}}2.365E-01 $^\circ$ \\ ±6.497E-02\end{tabular} & \begin{tabular}[c]{@{}l@{}}\textbf{2.073E-01} $\uparrow$ \\ ±3.804E-02\end{tabular} \\ \hline

D4        & \begin{tabular}[c]{@{}l@{}}2.926E-01    \\ ±3.910E-02\end{tabular} & \begin{tabular}[c]{@{}l@{}}2.796E-01 $^\circ$ \\ ±5.552E-02\end{tabular} & \begin{tabular}[c]{@{}l@{}}3.204E-01 $^\circ$ \\ ±1.278E-01\end{tabular} & \begin{tabular}[c]{@{}l@{}}\textbf{2.444E-01} $^\circ$ \\ ±4.279E-02\end{tabular} \\ \hline

D5      & \begin{tabular}[c]{@{}l@{}}1.681E-01    \\ ±4.306E-02\end{tabular} & \begin{tabular}[c]{@{}l@{}}9.710E-02 $\uparrow$ \\ ±6.325E-02\end{tabular} & \begin{tabular}[c]{@{}l@{}}1.174E-01 $^\circ$ \\ ±7.052E-02\end{tabular} & \begin{tabular}[c]{@{}l@{}}\textbf{5.507E-02} $\uparrow$ \\ ±4.262E-02\end{tabular} \\ \hline

D6           & \begin{tabular}[c]{@{}l@{}}2.737E-01    \\ ±2.179E-02\end{tabular} & \begin{tabular}[c]{@{}l@{}}\textbf{2.579E-01} $\uparrow$ \\ ±7.095E-02\end{tabular} & \begin{tabular}[c]{@{}l@{}}2.842E-01 $^\circ$ \\ ±6.715E-02\end{tabular} & \begin{tabular}[c]{@{}l@{}}2.649E-01 $^\circ$ \\ ±6.256E-02\end{tabular} \\ \hline

D7         & \begin{tabular}[c]{@{}l@{}}8.333E-02    \\ ±4.725E-02\end{tabular} & \begin{tabular}[c]{@{}l@{}}1.083E-01 $^\circ$ \\ ±6.798E-02\end{tabular} & \begin{tabular}[c]{@{}l@{}}\textbf{6.250E-02} $\uparrow$ \\ ±6.166E-02\end{tabular} & \begin{tabular}[c]{@{}l@{}}\textbf{7.639E-02} $^\circ$ \\ ±3.643E-02\end{tabular} \\ \hline

D8 & \begin{tabular}[c]{@{}l@{}}4.638E-02    \\ ±1.990E-02\end{tabular} & \begin{tabular}[c]{@{}l@{}}1.087E-01 $\downarrow$ \\ ±4.810E-02\end{tabular} & \begin{tabular}[c]{@{}l@{}}5.942E-02 $\uparrow$ \\ ±7.083E-02\end{tabular} & \begin{tabular}[c]{@{}l@{}}\textbf{3.478E-02} $^\circ$ \\ ±3.682E-02\end{tabular} \\ \hline

D9      & \begin{tabular}[c]{@{}l@{}}1.862E-01    \\ ±1.623E-02\end{tabular} & \begin{tabular}[c]{@{}l@{}}1.684E-01 $^\circ$  \\ ±3.060E-02\end{tabular}  & \begin{tabular}[c]{@{}l@{}}1.632E-01 $^\circ$ \\ ±3.095E-02\end{tabular} & \begin{tabular}[c]{@{}l@{}}\textbf{1.287E-01}  $\uparrow$ \\ ±2.062E-02\end{tabular}\\ \hline

D10      & \begin{tabular}[c]{@{}l@{}}2.086E-01    \\ ±1.666E-02\end{tabular} & \begin{tabular}[c]{@{}l@{}}1.624E-01 $\uparrow$ \\ ±3.333E-02\end{tabular} & \begin{tabular}[c]{@{}l@{}}1.839E-01 $^\circ$ \\ ±6.234E-02\end{tabular} & \begin{tabular}[c]{@{}l@{}}\textbf{1.312E-01} $\uparrow$ \\ ±3.274E-02\end{tabular} \\ \hline

D11      & \begin{tabular}[c]{@{}l@{}}3.505E-01    \\ ±2.951E-02\end{tabular} & \begin{tabular}[c]{@{}l@{}}2.848E-01 $\uparrow$\\  ±5.060E-02\end{tabular} & \begin{tabular}[c]{@{}l@{}}2.705E-01 $^\circ$ \\ ±3.951E-02\end{tabular} & \begin{tabular}[c]{@{}l@{}}\textbf{2.448E-01} $^\circ$ \\ ±5.181E-02\end{tabular} \\ \hline

D12       & \begin{tabular}[c]{@{}l@{}}2.207E-01    \\ ±1.623E-02\end{tabular} & \begin{tabular}[c]{@{}l@{}}1.839E-01 $\uparrow$ \\ ±4.558E-02\end{tabular} & \begin{tabular}[c]{@{}l@{}}1.828E-01 $^\circ$ \\ ±3.611E-02\end{tabular} & \begin{tabular}[c]{@{}l@{}}\textbf{1.483E-01} $\uparrow$ \\ ±2.625E-02\end{tabular} \\ \hline

D13          & \begin{tabular}[c]{@{}l@{}}1.492E-01    \\ ±8.049E-03\end{tabular} & \begin{tabular}[c]{@{}l@{}}9.894E-02 $\uparrow$ \\ ±2.593E-02\end{tabular} & \begin{tabular}[c]{@{}l@{}}7.778E-02 $\uparrow$\\  ±2.407E-02\end{tabular} & \begin{tabular}[c]{@{}l@{}}\textbf{6.138E-02}  $\uparrow$\\ ±2.757E-02\end{tabular} \\ \hline

D14           & \begin{tabular}[c]{@{}l@{}}9.487E-02    \\ ±1.986E-02\end{tabular} & \begin{tabular}[c]{@{}l@{}}1.141E-01 $^\circ$  \\ ±4.101E-02\end{tabular} & \begin{tabular}[c]{@{}l@{}}1.641E-01  $\downarrow$\\ ±2.660E-02\end{tabular} & \begin{tabular}[c]{@{}l@{}}\textbf{1.256E-01} $\uparrow$ \\ ±3.903E-02\end{tabular} \\ \hline

D15        & \begin{tabular}[c]{@{}l@{}}3.422E-01    \\ ±3.204E-02\end{tabular} & \begin{tabular}[c]{@{}l@{}}2.856E-01 $\uparrow$ \\ ±4.927E-02\end{tabular} & \begin{tabular}[c]{@{}l@{}}2.800E-01 $^\circ$ \\ ±6.349E-02\end{tabular} & \begin{tabular}[c]{@{}l@{}}\textbf{2.756E-01} $^\circ$ \\ ±5.099E-02\end{tabular} \\ \hline
R.       & 2.36 (15/0/0)                                                              & 2.17 (14/0/1)                                                           & 2.16  (13/0/2)                                                         & \textbf{1.44}                                                             \\ \hline
\end{tabular}
\end{table}

\section{Conclusions}

This paper presented LMSSS, a novel multi-objective evolutionary algorithm designed to tackle the challenges associated with large-scale feature selection. Our approach integrates several key strategies to enhance the efficiency and effectiveness of the evolutionary process in high-dimensional search spaces.
First, we introduced a shrinking scheme that reduces the dimensionality of the problem by eliminating irrelevant features before the main evolutionary process. This is achieved through a ranking-based filtering method that evaluates features based on their correlation with class labels and their frequency in an initial, cost-effective evolutionary process.
Second, a smart crossover and mutation scheme based on voting between parent solutions was developed. The crossover mechanism makes decisions about each feature based on the agreement or disagreement between the parents, with a higher weight assigned to the parent demonstrating better classification accuracy. This approach helps in generating more diverse and potentially optimal feature subsets.
The intelligent mutation process that targets features prematurely excluded from the population. This ensures these features are given another opportunity to be evaluated in combination with other features, thereby enhancing the thoroughness of the feature selection process.

We evaluated LMSSS on fifteen well-known large-scale datasets and compared it with seven state-of-the-art and recent large-scale feature selection algorithms. The experimental results demonstrate the effectiveness of the proposed LMSSS algorithm in handling complex, high-dimensional datasets in terms of multi-objective evaluation metrics and feature selection metrics. The integration of the proposed techniques not only reduces computational costs but also improves model performance by selecting the most relevant features. The shrinking process efficiently reduces the search space, alleviating the challenges of exploring large-scale spaces. By prioritizing the most relevant features early in the process, our method prevents the optimizer from wasting time and computational power on the least relevant features. Moreover, assigning greater weight to the accuracy of classification, a more challenging objective in imbalanced feature selection optimization problems, enables the optimizer to identify feature sets with higher accuracy. This shrinking technique  not only leads to higher-quality feature subsets but also enhances computational efficiency without compromising performance.

Our future work will concentrate on further optimizing the algorithm and extending its application to a broader array of real-world problems. The shrinking technique holds promise as a general framework for many large-scale optimization problems. Exploring criteria to filter out less prominent variables from the search space will be a key focus.


\bibliographystyle{elsarticle-num} 
\bibliography{main.bib}



\end{document}